\documentclass[10pt,journal,compsoc]{IEEEtran}
\ifCLASSOPTIONcompsoc
  \usepackage[nocompress]{cite}
\else
  \usepackage{cite}
\fi
\ifCLASSINFOpdf
\else
\fi

\usepackage{multirow}
\usepackage{xcolor}
\usepackage{graphicx}
\usepackage{amsmath}
\usepackage{amssymb}
\usepackage{wrapfig}
\usepackage{diagbox}
\usepackage{booktabs}
\usepackage{color}
\usepackage{verbatim}
\usepackage{caption}
\usepackage{ulem}
\usepackage{subfig}
\usepackage{mathrsfs}
\usepackage[pagebackref=false,breaklinks=true,colorlinks,bookmarks=false,urlcolor=blue]{hyperref}
\newcommand{\tabincell}[2]{\begin{tabular}{@{}#1@{}}#2\end{tabular}}

\newcommand{\Gary}[1]{\textcolor{black}{#1}}
\newcommand{\why}[1]{\textcolor{black}{#1}}
\newcommand{\qy}[1]{\textcolor{black}{#1}}
\newcommand{\hao}[1]{\textcolor{black}{#1}}

\begin{document}
%
\title{Deep Learning for 3D Point Clouds: A Survey}
%
%
%
%

\author{Yulan Guo\textsuperscript{$\ast$},
        Hanyun Wang\textsuperscript{$\ast$},
        Qingyong Hu\textsuperscript{$\ast$},
        Hao Liu\textsuperscript{$\ast$},
        Li Liu,
        and Mohammed Bennamoun
\IEEEcompsocitemizethanks{\IEEEcompsocthanksitem Y Guo and H. Liu are with the School of Electronics and Communication Engineering, Sun Yat-sen University, China. H. Wang is with the School of Surveying and Mapping, Information Engineering University, China. Q. Hu is with Department of Computer Science, University of Oxford, UK. 
L. Liu is with the College of System Engineering, National University of Defense Technology, China, and also with the Center for Machine Vision and Signal Analysis, University of Oulu, Finland.
M. Bennamoun is with the Department of Computer Science and Software Engineering, the University of Western Australia, Australia. 
\IEEEcompsocthanksitem *Y. Guo, H. Wang, Q. Hu and H. Liu have equal contribution to this work and are co-first authors.
\IEEEcompsocthanksitem Corresponding author: Yulan Guo (yulan.guo@nudt.edu.cn).
}
}

%
%

\markboth{IEEE Transactions on pattern analysis and machine intelligence}
{Guo \MakeLowercase{\textit{et al.}}: Deep Learning for 3D Point Clouds: A Survey}
%



\IEEEtitleabstractindextext{%
\begin{abstract}
Point cloud learning has lately attracted increasing attention due to its wide applications in many areas, such as computer vision, autonomous driving, and robotics. As a dominating technique in AI, deep learning has been successfully used to solve various 2D vision problems. However, deep learning on point clouds is still in its infancy due to the unique challenges faced by the processing of point clouds with deep neural networks. Recently, deep learning on point clouds has become even thriving, with numerous methods being proposed to address different problems in this area. To stimulate future research, this paper presents a comprehensive review of recent progress in deep learning methods for point clouds. It covers three major tasks, including 3D  shape  classification, 3D  object  detection and tracking, and 3D point cloud segmentation. It also presents  comparative results on several publicly available datasets, together with insightful observations and inspiring future research directions. 
\end{abstract}

}

\maketitle

\IEEEdisplaynontitleabstractindextext

%
\IEEEpeerreviewmaketitle

\IEEEraisesectionheading{\section{Introduction}\label{sec:introduction}}

%
%
%
%
\IEEEPARstart{W}{ith} the rapid development of 3D acquisition technologies, 3D sensors are becoming increasingly available and affordable, including various types of 3D scanners, LiDARs, and RGB-D cameras (such as Kinect, RealSense and Apple depth cameras) \cite{liang2019stereo}. 3D data acquired by these sensors can provide rich geometric, shape and scale information \cite{guo2013rotational,guo2014Object}. Complemented with 2D images, 3D data provides an opportunity for a better understanding of the surrounding environment for machines. 3D data has numerous applications in different areas, including autonomous driving, robotics, remote sensing, and medical treatment \cite{chen2017multi}. 

3D data can usually be represented with different formats, including depth images, point clouds, meshes, and volumetric grids. As a commonly used format, point cloud representation preserves the original geometric information in 3D space without any discretization. Therefore, it is the preferred representation for many scene understanding related applications such as autonomous driving and robotics. Recently, deep learning techniques have dominated many research areas, such as computer vision, speech recognition, and natural language processing. However, deep learning on 3D point clouds still face several significant challenges \cite{PointNet}, such as the small scale of datasets, the high dimensionality and the unstructured nature of 3D point clouds. On this basis, this paper focuses on the analysis of deep learning methods which have been used to process 3D point clouds.

\begin{figure*}[t]
\centering
\includegraphics[width=\textwidth]{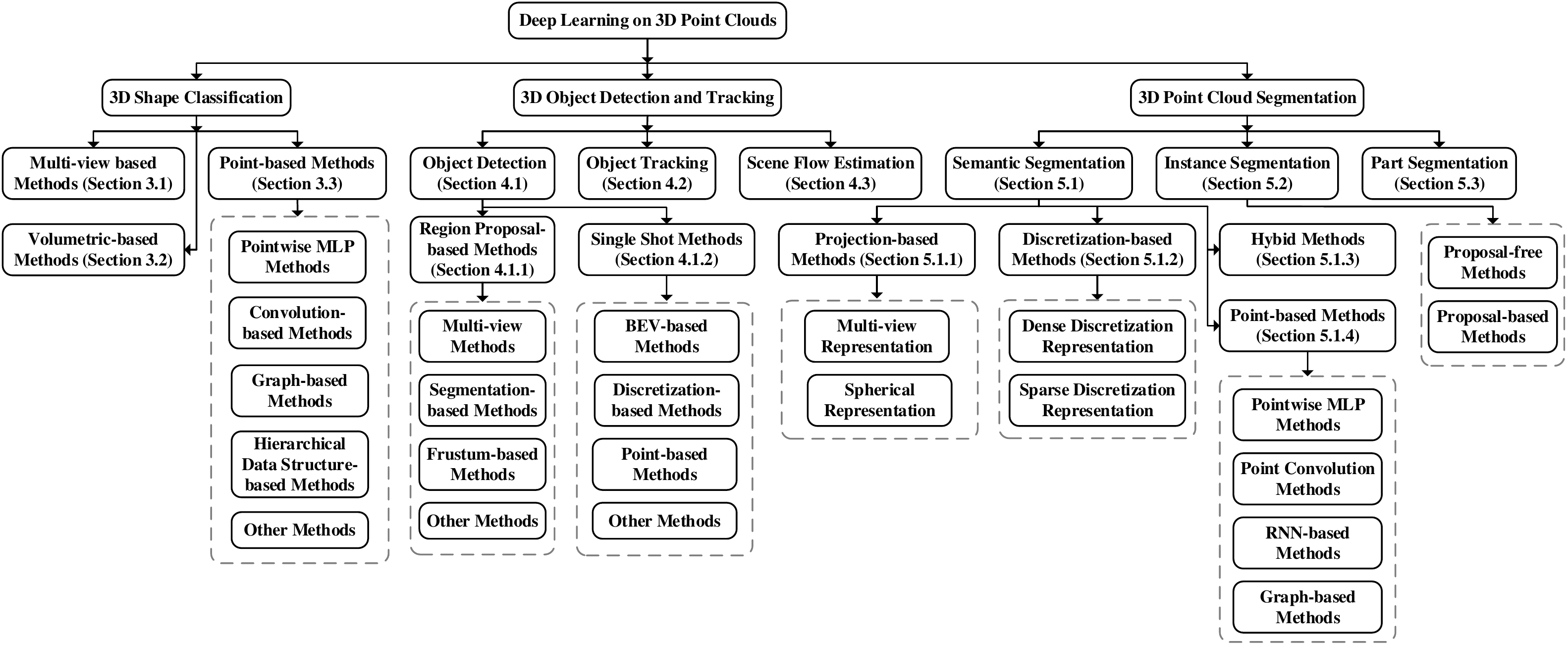}
\caption{A taxonomy of deep learning methods for 3D point clouds.
\label{fig:structure}}
\end{figure*}

Deep learning on point clouds has been attracting more and more attention, especially in the last five years. Several publicly available datasets are also released, such as ModelNet \cite{wu20153d}, ScanObjectNN \cite{ScanObjecNN}, ShapeNet \cite{Chang2015ShapeNet}, PartNet \cite{Partnet}, S3DIS \cite{S3DIS}, ScanNet \cite{ScanNet}, Semantic3D \cite{Semantic3d}, ApolloCar3D \cite{song2019apollocar3d}, and the KITTI Vision Benchmark Suite \cite{KITTI, semantickitti}. These datasets have further boosted the research of deep learning on 3D point clouds, with an increasingly number of methods being proposed to address various problems related to point cloud processing, including 3D shape classification, 3D object detection and tracking,  \Gary{3D point cloud segmentation, 3D point cloud registration, 6-DOF pose estimation, and 3D reconstruction \cite{elbaz20173d,zeng2017multi,han2019image}}. Few surveys of deep learning on 3D data are also available, such as \cite{ioannidou2017deep,ahmed2018deep,Xie2019review,rahman2019recent}. \Gary{However, our paper is the first to specifically focus on deep learning methods for point cloud understanding.} 
A taxonomy of existing deep learning methods for 3D point clouds \why{is} shown in Fig. \ref{fig:structure}. 

Compared with the existing literatures, the major contributions of this work can be summarized as follows:
\begin{enumerate}
    \item To the best of our knowledge, this is the \textit{first} survey paper to comprehensively cover deep learning  methods for several important point cloud understanding tasks, including 3D shape classification, 3D object detection and tracking, and 3D point cloud segmentation.
    \item As opposed to existing reviews \cite{ahmed2018deep,ioannidou2017deep}, we specifically focus on deep learning methods for \textit{3D point clouds} rather than all types of 3D data. 
    \item This paper covers the \textit{most recent and advanced progresses} of deep learning on point clouds. Therefore, it provides the readers with the state-of-the-art methods.
    \item Comprehensive \textit{comparisons of existing methods} on several publicly available datasets are provided (e.g., in Tables \ref{Tab:ModelNet}, \ref{Tab:KITTI3D}, \ref{Tab:KITTIBEV}, \ref{tab:segmentation_results}), with brief summaries and insightful discussions being presented. 
\end{enumerate}

\begin{table*}[t]
\caption{A summary of existing datasets for 3D shape classification, 3D object detection and tracking, and 3D point cloud segmentation. \protect\textsuperscript{1} The number of classes used for evaluation and the number of annotated classes (shown in brackets).\label{tab:Overview-dataset}}
\begin{tabular}{|r|c|c|c|c|c|c|c|}
\hline 
\multicolumn{8}{|c|}{\textbf{Datasets for 3D Shape Classification}}\\
\hline 
Name and Reference & Year & \#Samples & \#Classes & \#Training & \#Test & Type & Representation\\
\hline 
McGill Benchmark\cite{Siddiqi2008Retrieving}  & 2008 & 456 & 19 & 304 & 152 & Synthetic & Mesh\\
\hline 
Sydney Urban Objects\cite{SYDNEY-URBAN-OBJECTS-DATASET}  & 2013 & 588 & 14 & - & - & Real-World & Point Clouds\\
\hline 
ModelNet10\cite{wu20153d}  & 2015 & 4899 & 10 & 3991 & 605 & Synthetic & Mesh\\
\hline 
ModelNet40\cite{wu20153d}  & 2015 & 12311 & 40 & 9843 & 2468 & Synthetic & Mesh\\
\hline 
ShapeNet\cite{Chang2015ShapeNet}  & 2015 & 51190 & 55 & - & - & Synthetic & Mesh\\
\hline 
ScanNet\cite{ScanNet}  & 2017 & 12283 & 17 & 9677 & 2606 & Real-World & RGB-D\\
\hline 
ScanObjectNN\cite{ScanObjecNN}  & 2019 & 2902 & 15 & 2321 & 581 & Real-World & Point Clouds\\
\hline 
\multicolumn{8}{|c|}{\textbf{Datasets for 3D Object Detection and Tracking}}\\
\hline 
Name and Reference & Year & \#Scenes & \#Classes & \#Annotated Frames & \#3D Boxes & Secne Type & Sensors\\
\hline 
KITTI \cite{KITTI} & 2012  & 22  & 8  & 15K  & 200K  & Urban (Driving)  & RGB \& LiDAR \\
\hline 
SUN RGB-D \cite{SunRGBD} & 2015  & 47  & 37  & 5K  & 65K  & Indoor  & RGB-D \\
\hline 
ScanNetV2 \cite{ScanNet} & 2018  & 1.5K  & 18  & -  & -  & Indoor  & RGB-D \& Mesh \\
\hline 
H3D \cite{H3D} & 2019  & 160  & 8  & 27K  & 1.1M  & Urban (Driving)  & RGB \& LiDAR \\
\hline 
Argoverse \cite{Argoverse} & 2019  & 113  & 15  & 44K  & 993K  & Urban (Driving)  & RGB \& LiDAR \\
\hline 
Lyft L5 \cite{Lyft} & 2019  & 366  & 9  & 46K  & 1.3M  & Urban (Driving)  & RGB \& LiDAR \\
\hline 
A{*}3D \cite{A*3D} & 2019  & -  & 7  & 39K  & 230K  & Urban (Driving)  & RGB \& LiDAR \\
\hline 
Waymo Open \cite{Waymo} & 2020  & 1K  & 4  & 200K  & 12M  & Urban (Driving)  & RGB \& LiDAR \\
\hline 
nuScenes \cite{nuScenes} & 2020  & 1K  & 23  & 40K  & 1.4M  & Urban (Driving)  & RGB \& LiDAR \\
\hline 
\multicolumn{8}{|c|}{\textbf{Datasets for 3D Point Cloud Segmentation}}\\
\hline 
Name and Reference & Year & \#Points & \#Classes\textsuperscript{1} & \#Scans & Spatial Size & RGB & Sensors\\
\hline 
Oakland\cite{oakland} & 2009 & 1.6M & 5(44) & 17 & - & N/A & MLS\\
\hline 
ISPRS\cite{rottensteiner2012isprs} & 2012 & 1.2M & 9 & - & - & N/A &  ALS\\
\hline 
Paris-rue-Madame\cite{paris-rue-madame} & 2014 & 20M & 17 & 2 & - & N/A & MLS\\
\hline 
IQmulus\cite{IQmulus} & 2015 & 300M & 8(22) & 10 & - & N/A & MLS\\
\hline 
ScanNet\cite{ScanNet} & 2017 & - & 20(20) & 1513 & 8$\times$4$\times$4 & Yes & RGB-D\\
\hline 
S3DIS\cite{S3DIS} & 2017 & 273M & 13(13) & 272 & 10$\times$5$\times$5 & Yes & Matterport\\
\hline 
Semantic3D\cite{Semantic3d} & 2017 & 4000M & 8(9) & 15/15 & 250$\times$260$\times$80 & Yes &  TLS\\
\hline 
Paris-Lille-3D\cite{Paris_Lille_3D} & 2018 & 143M & 9(50) & 3 & 200$\times$280$\times$ 30 & N/A & MLS\\
\hline 
SemanticKITTI\cite{semantickitti} & 2019 & 4549M & 25(28) & 23201/20351 & 150$\times$100$\times$10 & N/A & MLS\\
\hline 
Toronto-3D\cite{Toronto3D} & 2020 & 78.3M & 8(9) & 4 & 260$\times$350$\times$ 40 & Yes & MLS\\
\hline 
DALES\cite{varney2020dales} & 2020 & 505M & 8(9) & 40 & 500$\times$500$\times$65 & N/A &  ALS\\
\hline 
\end{tabular}
\end{table*}

The structure of this paper is as follows. \Gary{Section \ref{sec:background} introduces the datasets and evaluation metrics for the respective tasks.} 
Section \ref{sec:shape_classification} reviews the methods for 3D shape classification. Section \ref{sec:object_detection} provides a survey of existing methods for 3D object detection and tracking. Section \ref{sec:scene_segmentation}  presents a review of methods for point cloud segmentation, including semantic segmentation, instance segmentation, and part segmentation. Finally, Section \ref{sec:conclusion} concludes the paper. We also provide a regularly updated project page on: \url{https://github.com/QingyongHu/SoTA-Point-Cloud}.

\qy{\section{Background}\label{sec:background}}

\qy{\subsection{Datasets}}
\why{A large number of datasets have been collected to evaluate the performance of deep learning algorithms for different 3D point clouds applications. Table \ref{tab:Overview-dataset} lists some typical datasets used for 3D shape classification, \qy{3D object detection and tracking, and 3D point cloud segmentation}. In particular, the attributes of these datasets are also summarized.}

\why{For 3D shape classification, there are two types of datasets: synthetic datasets \cite{wu20153d, Chang2015ShapeNet} and real-world datasets \cite{ScanNet, ScanObjecNN}. Objects in the synthetic datasets are complete, without any occlusion and background. In contrast, objects in the real-world datasets are occluded at different levels and some objects are contaminated with background noise.}

\hao{For 3D object detection and tracking, there are two types of datasets: indoor scenes \cite{SunRGBD,ScanNet} and outdoor urban scenes \cite{KITTI,Lyft,nuScenes,Waymo}. The point clouds in the indoor datasets are either converted from dense depth maps or sampled from 3D  meshes. The outdoor urban datasets are designed for autonomous driving, where objects are spatially well separated and these point clouds are sparse.}

\qy{For 3D point cloud segmentation, these datasets are acquired by different types of sensors, including Mobile Laser Scanners (MLS) \cite{semantickitti, paris-rue-madame, Paris_Lille_3D}, Aerial Laser Scanners (ALS) \cite{rottensteiner2012isprs, varney2020dales}, static Terrestrial Laser Scanners (TLS) \cite{Semantic3d}, RGB-D cameras \cite{ScanNet} and other 3D scanners \cite{S3DIS}.
These datasets can be used to develop algorithms for various challenges including similar distractors, shape incompleteness, and class imbalance.}

\qy{\subsection{Evaluation Metrics}}
\qy{Different evaluation metrics have been proposed to test these methods for various point cloud understanding tasks.} \why{For 3D shape classification, \textit{Overall Accuracy} (OA) and \textit{mean class accuracy} (mAcc) are the most frequently used performance criteria. `OA' represents the mean accuracy for all test instances and `mAcc' represents the mean accuracy for all shape classes}. 
\hao{For 3D object detection, \textit{Average Precision} (AP) is the most frequently used criterion. It is calculated as the area under the precision-recall curve. \textit{Precision} and \textit{Success} are commonly used to evaluate the overall performance of a 3D single object tracker. \textit{Average Multi-Object Tracking Accuracy} (AMOTA) and \textit{Average Multi-Object Tracking Precision} (AMOTP) are the most frequently used criteria for the evaluation of 3D multi-object tracking.}
\qy{For 3D point cloud segmentation, OA, \textit{mean Intersection over Union} (mIoU) and \textit{mean class Accuracy} (mAcc) \cite{S3DIS, Semantic3d, Paris_Lille_3D, semantickitti, Toronto3D} are the most frequently used criteria for performance evaluation. In particular, \textit{mean Average Precision} (mAP) \cite{Scanet} is also used in instance segmentation of 3D point clouds.}

\section{3D Shape Classification}\label{sec:shape_classification}

\begin{figure*}[t]
\centering
\includegraphics[width=\textwidth]{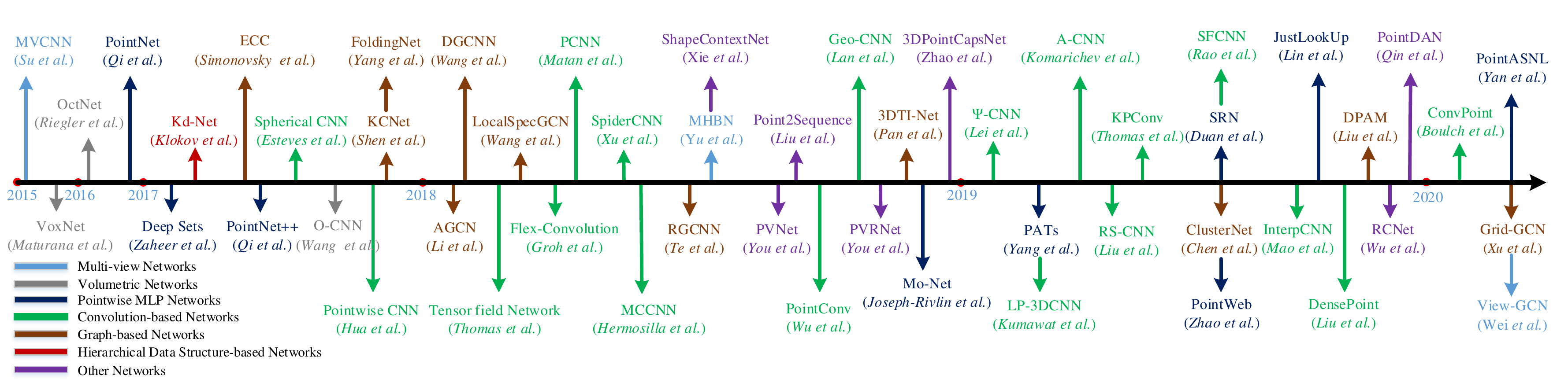}
\caption{\why{Chronological overview of the most relevant deep learning-based 3D shape classification methods.}
\label{fig:milestone_classification}}
\end{figure*}

Methods for this task usually learn the embedding of each point first and then extract a global shape embedding from the whole point cloud using an aggregation method. Classification is finally achieved by feeding the global embedding into several fully connected layers. According to the data type of input for neural networks, existing 3D shape classification methods can be divided into \why{multi-view based, volumetric-based} and point-based methods. Several milestone methods are illustrated in Fig. \ref{fig:milestone_classification}. 

\why{Multi-view based methods project an unstructured point cloud into 2D images, while volumetric-based methods convert a point cloud into a 3D volumetric representation. Then,  well-established 2D or 3D convolutional networks are leveraged to achieve shape classification.} In contrast, point-based methods directly work on raw point clouds without any voxelization or projection. Point-based methods do not introduce explicit information loss and become increasingly popular. Note that, this paper mainly focuses on point-based methods, but also includes few \why{multi-view based and volumetric-based} methods for completeness.

\subsection{\why{Multi-view based Methods}}

\why{These methods first project a 3D shape into multiple views and extract   view-wise features, and then fuse these features for accurate shape classification. How to aggregate multiple view-wise features into a discriminative global representation is a key challenge for these methods.}

MVCNN \cite{MVCNN} is a pioneering work, which simply max-pools multi-view features into a global descriptor. However, max-pooling only retains the maximum elements from a specific view, resulting in information loss. MHBN \cite{MHBN} integrates local convolutional features by harmonized bilinear pooling to produce a compact global descriptor. Yang et al. \cite{LRMV} first leveraged a relation network to exploit the inter-relationships (e.g., region-region relationship and view-view relationship) over a group of views, and then aggregated these views to obtain a discriminative 3D object representation. In addition, several other methods \cite{qi2016volumetric,GVCNN,wang2019dominant,ma2018learning} have also been proposed to improve the recognition accuracy. \why{Unlike previous methods, Wei et al. \cite{Wei_2020_CVPR} used a directed graph in View-GCN by considering multiple views as grpah nodes. The core layer composing of local graph convolution, non-local message passing and selective view-sampling is then applied to the constructed graph. The concatenation of max-pooled node features at all levels is finally used to form the global shape descriptor.}

\subsection{\why{Volumetric-based Methods}}
\why{These methods usually voxelize a point cloud into 3D grids, and then apply a 3D Convolution Neural Network (CNN) on the volumetric representation for shape classification.}

 \why{Maturana et al.} \cite{maturana2015voxnet} introduced a volumetric occupancy network called VoxNet to achieve robust 3D object recognition. Wu et al. \cite{wu20153d} proposed a convolutional deep belief-based 3D ShapeNets to learn the distribution of points from various 3D shapes (which are represented by a probability distribution of binary variables on voxel grids). Although encouraging performance has been achieved, these methods are unable to scale well to dense 3D data since the computation and memory footprint grow cubically with the resolution. 

To this end, a hierarchical and compact structure (such as octree) is introduced to reduce the computational and memory costs of these methods. OctNet \cite{Riegler2017} first hierarchically partitions a point cloud using a hybrid grid-octree structure, which represents the scene with several shallow octrees along a regular grid. The structure of octree is encoded efficiently using a bit string representation, and the feature vector of each voxel is indexed by simple arithmetic. Wang et al. \cite{O-cnn} proposed an Octree-based CNN for 3D shape classification. The average normal vectors of a 3D model sampled in the finest leaf octants are fed into the network, and 3D-CNN is applied on the octants occupied by the 3D shape surface. Compared to a baseline network based on dense input grids, OctNet requires much less memory and runtime for high-resolution point clouds. Le et al. \cite{le2018pointgrid} proposed a hybrid network called PointGrid,  which integrates the point and grid representation for efficient point cloud processing. A constant number of points is sampled within each embedding volumetric grid cell, which allows the network to extract geometric details by using 3D convolutions. Ben-Shabat et al. \cite{Ben-Shabat2017} transformed the input point cloud into 3D grids which are further represented by 3D modified Fisher Vector (3DmFV) method, and then learned the global representation through a conventional CNN architecture.

\subsection{Point-based Methods}
According to the network architecture used for the feature learning of each point, methods in this category can be divided into pointwise MLP, convolution-based, graph-based, \why{hierarchical data structure-based methods} and other typical methods.

\subsubsection{\bf Pointwise MLP Methods}

\begin{figure}[b]
\centering
\includegraphics[scale=0.65]{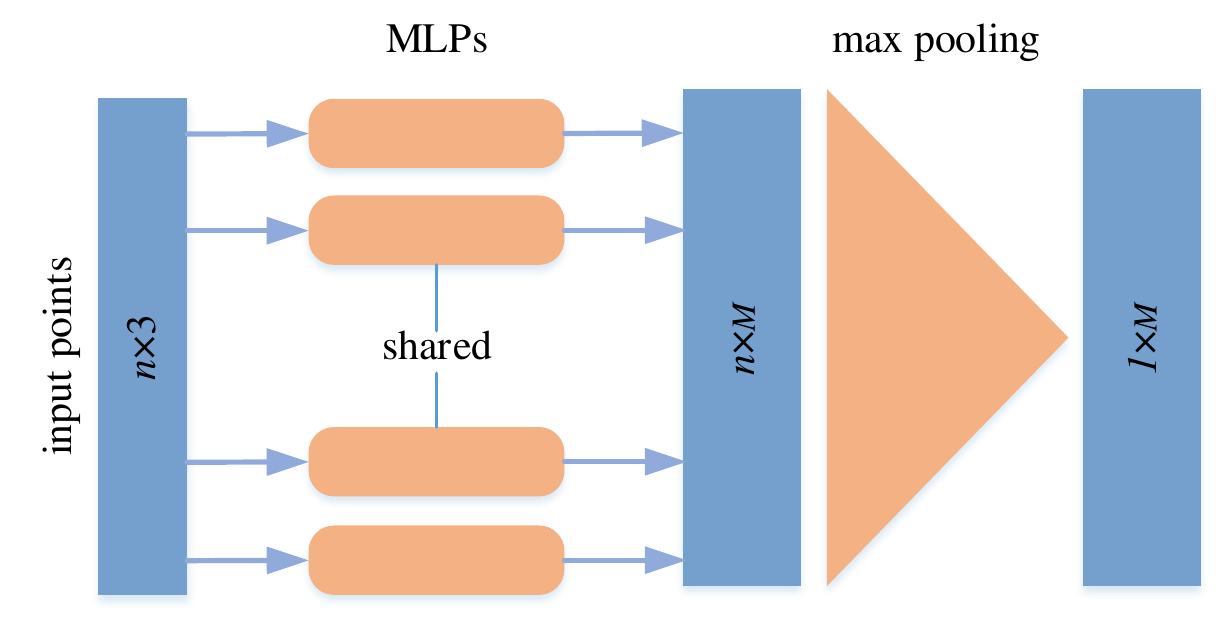}
\caption{\why{A lightweight architecture of PointNet}. $n$ denotes the number of input points, $M$ denotes the dimension of the learned features for each point.}
\label{fig:pointnet}
\end{figure}

These methods model each point independently with several shared Multi-Layer Perceptrons (MLPs) and then aggregate a global feature using a symmetric aggregation function, as shown in Fig. \ref{fig:pointnet}.

Typical deep learning methods for 2D images cannot be directly applied to 3D point clouds due to their inherent data irregularities. As a pioneering work, PointNet \cite{PointNet} directly takes point clouds as its input and achieves permutation invariance with a symmetric function. Specifically, PointNet learns pointwise features independently with several MLP layers and extracts global features with a max-pooling layer. Deep sets \cite{Zaheer2017} achieves permutation invariance by summing up all representations and applying nonlinear transformations. Since features are learned independently for each point in PointNet \cite{PointNet}, the local structural information between points cannot be captured. Therefore, Qi et al. \cite{PointNet++} proposed a hierarchical network PointNet++ to capture fine geometric structures from the neighborhood of each point. As the core of PointNet++ hierarchy, its set abstraction level is composed of three layers: the sampling layer, the grouping layer and the PointNet based learning layer. By stacking several set abstraction levels, PointNet++ learns features from a local geometric structure and abstracts the local features layer by layer.

Because of its simplicity and strong representation ability, many networks have been developed based on PointNet \cite{PointNet}. The architecture of Mo-Net \cite{Joseph-Rivlin2018} is similar to PointNet \cite{PointNet} but it takes a finite set of moments as its input. Point Attention Transformers (PATs) \cite{Yang2019} represents each point by its own absolute position and relative positions with respect to its neighbors and learns high dimensional features through MLPs. Then, Group Shuffle Attention (GSA) is used to capture relations between points, and a permutation invariant, differentiable and trainable end-to-end Gumbel Subset Sampling (GSS) layer is developed to learn hierarchical features. Based on PointNet++ \cite{PointNet++}, PointWeb \cite{PointWeb2019} utilizes the context of the local neighborhood to improve point features using Adaptive Feature Adjustment (AFA). Duan et al. \cite {Duan2019} proposed a Structural Relational Network (SRN) to learn structural relational features between different local structures using MLP. Lin et al. \cite{Lin2019Justlookup} accelerated the inference process by constructing a lookup table for both input and function spaces learned by PointNet. The inference time on the ModelNet and ShapeNet datasets is sped up by 1.5 ms and 32 times over PointNet on a moderate machine. SRINet \cite{Sun2019SRINet} first projects a point cloud to obtain rotation invariant representations, and then utilizes PointNet-based backbone to extract a global feature and graph-based aggregation to extract local features. \why{In PointASNL, Yan et al. \cite{Yan_2020_CVPR} utilized an Adaptive Sampling (AS) module to adaptively adjust the coordinates and features of points sampled by the Furthest Point Sampling (FPS) algorithm, and proposed a local-non-local (L-NL) module to capture the local and long range dependencies of these sampled points.}

\subsubsection{\bf Convolution-based Methods}
\begin{figure}[h]
\centering
\includegraphics[width=\columnwidth,keepaspectratio]{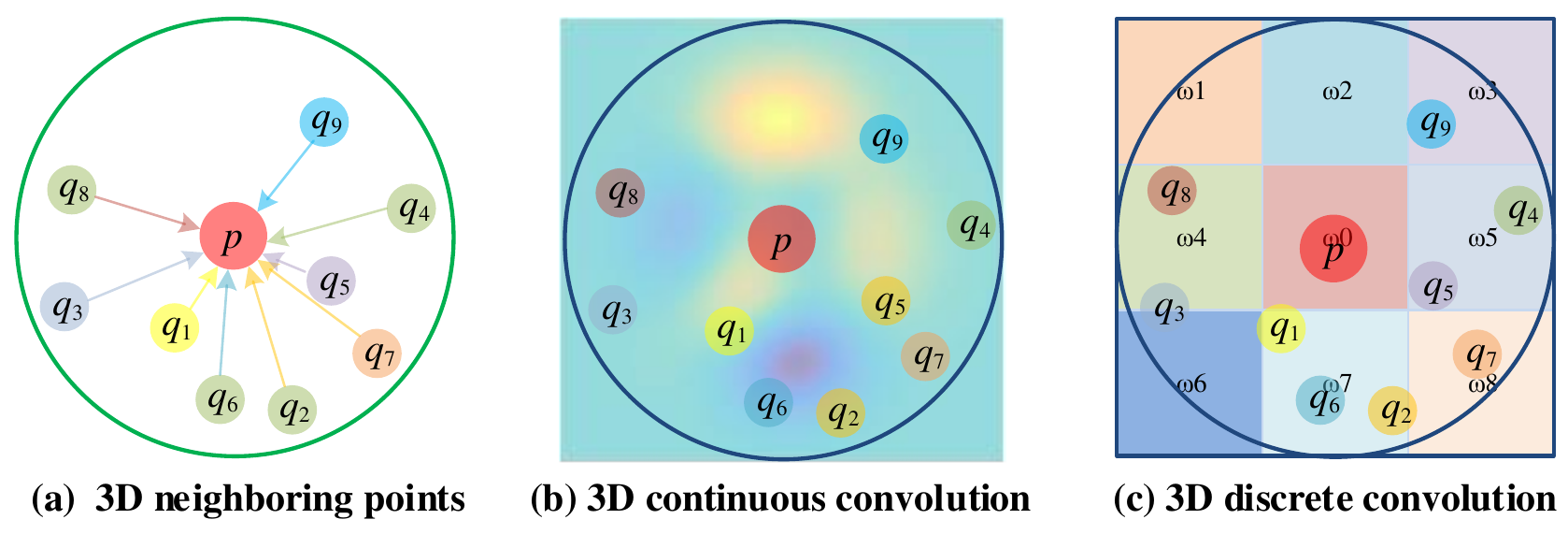}
\caption{\label{fig:3D Convolution Network} An illustration of a continuous and discrete convolution for local neighbors of a point. (a) represents a local neighborhood ${q_i}$ centered at point $p$; (b) and (c) represent 3D continuous and discrete convolution, respectively.}
\end{figure}

Compared with kernels defined on 2D grid structures (e.g., images), convolutional kernels for 3D point clouds are hard to design due to the irregularity of point clouds. According to the type of convolutional kernels, current 3D convolution methods can be divided into continuous and discrete convolution methods, as shown in Fig. \ref{fig:3D Convolution Network}.

\textbf{3D Continuous Convolution Methods.} 
These methods define convolutional kernels on a continuous space, where the weights for neighboring points are related to the spatial distribution with respect to the center point.

3D convolution can be interpreted as a weighted sum over a given subset. As the core layer of RS-CNN \cite{Liu2019}, RS-Conv takes a local subset of points around a certain point as its input, and the convolution is implemented using an MLP by learning the mapping from low-level relations (such as Euclidean distance and relative position) to high-level relations between points in the local subset.  In \cite{Boulch2019}, kernel elements are selected randomly in a unit sphere. An MLP-based continuous function is then used to establish relation between the locations of the kernel elements and the point cloud. In DensePoint \cite{Densepoint2019}, convolution is defined as a Single-Layer Perceptron (SLP) with a nonlinear activator. Features are learned by concatenating features from all previous layers to sufficiently exploit the contextual information. Thomas et al. \cite{kpconv} proposed both rigid and deformable Kernel Point Convolution (KPConv) operators for 3D point clouds using a set of learnable kernel points. ConvPoint \cite{Boulch2019ConvPoint} separates the convolution kernel into spatial and feature parts. The locations of the spatial part are randomly selected from a unit sphere and the weighting function is learned through a simple MLP.

Some methods also use existing algorithms to perform convolution. In PointConv \cite{Wu2018}, convolution is defined as a Monte Carlo estimation of the continuous 3D convolution with respect to an importance sampling. The convolutional kernels consist of a weighting function (which is learned with MLP layers) and a density function (which is learned by a kernelized density estimation and an MLP layer). To improve memory and computational efficiency, the 3D convolution is further reduced into two operations: matrix multiplication and 2D convolution. With the same parameter setting, its memory consumption can be reduced by about 64 times. In MCCNN \cite{Hermosilla2018}, convolution is  considered as a Monte Carlo estimation process relying on a sample's density function (which is implemented with MLP). Poisson disk sampling is then used to construct a point cloud hierarchy. This convolution operator can be used to perform convolution between two or multiple sampling methods and can handle varying sampling densities. In SpiderCNN \cite{Xu2018}, SpiderConv is proposed to define convolution as the product of a step function and a Taylor expansion defined on the $k$ nearest neighbors. The step function captures the coarse geometry by encoding the local geodesic distance, and the Taylor expansion captures the intrinsic local geometric variations by interpolating arbitrary values at the vertices of a cube. Besides, a convolution network PCNN \cite{Matan2018} is also proposed for 3D point clouds based on the radial basis function. 

Several methods have been proposed to address the rotation equivariant problem faced by 3D convolution networks. Esteves et al. \cite{Esteves2017} proposed 3D Spherical CNN to learn rotation equivariant representation for 3D shapes, which takes multi-valued spherical functions as its input. Localized convolutional filters are obtained by parameterizing spectrum with anchor points in the spherical harmonic domain. Tensor field networks \cite{Thomas2018} are proposed to define the point convolution operation as the product of a learnable radial function and  spherical harmonics, which are locally equivariant to 3D rotations, translations, and permutations. The convolution in \cite{Cohen2018} is defined based on the spherical cross-correlation and implemented using a generalized Fast Fourier Transformation (FFT) algorithm. Based on PCNN, SPHNet \cite{Poulenard2019SPHNet} achieves rotation invariance by incorporating spherical harmonic kernels during convolution on volumetric functions.  

To accelerate computing speed, Flex-Convolution \cite{Flexconv} defines weights of convolution kernel as standard scalar product over $k$ nearest neighbors, which can be accelerated using CUDA. Experimental results have demonstrated its competitive performance on a small dataset with fewer parameters and lower memory consumption.

\textbf{3D Discrete Convolution Methods.} \label{3DGN}
These methods define convolutional kernels on regular grids, where the weights for neighboring points are related to the offsets with respect to the center point.

Hua et al. \cite{Hua2018} transformed non-uniform 3D point clouds into uniform grids and defined convolutional kernels on each grid. The proposed 3D kernel assigns the same weights to all points falling into the same grid. For a given point, the mean features of all the neighboring points that are located on the same grid are computed from the previous layer. Then, mean features of all grids are weighted and summed to produce the output of the current layer. Lei et al. \cite{lei2019octree} defined a spherical convolutional kernel by partitioning a 3D spherical neighboring region into multiple volumetric bins and associating each bin with a learnable weighting matrix. The output of the spherical convolutional kernel for a point is determined by the non-linear activation of the mean of weighted activation values of its neighboring points. In GeoConv \cite{Lan2018}, the geometric relationship between a point and its neighboring points is explicitly modeled based on six bases. Edge features along each direction of the basis are weighted independently by a direction-associated learnable matrix. These direction-associated features are then aggregated according to the angles formed by the given point and its neighboring points. For a given point, its feature at the current layer is defined as the sum of features of the given point and its neighboring edge features at the previous layer. 

PointCNN \cite{li2018pointcnn} transforms the input points into a latent and potentially canonical order through a $\chi$-conv transformation (which is implemented through MLP) and then applies typical convolutional operator on the transformed features. By interpolating point features to neighboring discrete convolutional kernel-weight coordinates, Mao et al. \cite{Mao2019InterpConv} proposed an interpolated convolution operator InterpConv to measure the geometric relations between input point clouds and kernel-weight coordinates. Zhang et al. \cite{Zhang2019RIConv} proposed a RIConv operator to achieve rotation invariance, which takes low-level rotation invariant geometric features as input and then turns the convolution into 1D by a simple binning approach. A-CNN \cite{Komarichev2019} defines an annular convolution by looping the array of neighbors with respect to the size of kernel on each ring of the query point and learns the relationship between neighboring points in a local subset. 

To reduce the computational and memory cost of 3D CNNs, Kumawat et al. \cite{Kumawat2019} proposed a Rectified Local Phase Volume (ReLPV) block to extract phase in a 3D local neighborhood based on 3D Short Term Fourier Transform (STFT), which significantly reduces the number of parameters. In SFCNN \cite{SFCNN2019}, a point cloud is projected onto regular icosahedral lattices with aligned spherical coordinates. Convolutions are then conducted upon the features concatenated from vertices of spherical lattices and their neighbors through convolution-maxpooling-convolution structures. SFCNN is resistant to rotations and perturbations.

\subsubsection{\bf Graph-based Methods}

\begin{figure}[h]
\centering
\includegraphics[width=\columnwidth,keepaspectratio]{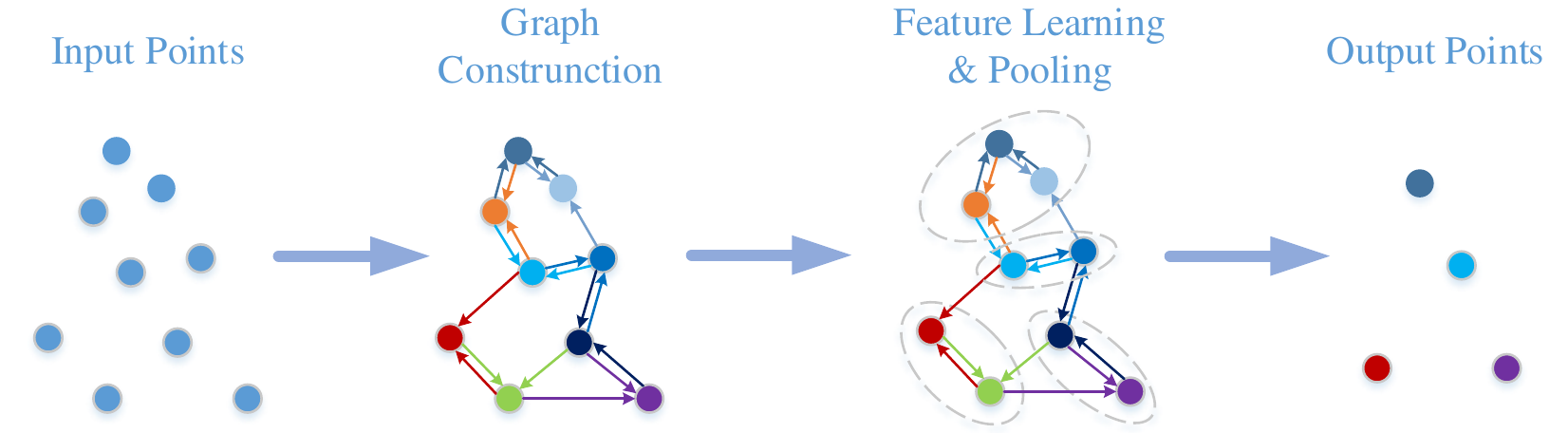}
\caption{An illustration of a graph-based network.}
\label{fig:Graph_based_Network}
\end{figure}

Graph-based networks consider each point in a point cloud as a vertex of a graph, and generate directed edges for the graph based on the neighbors of each point. Feature learning is then performed in spatial or spectral domains \cite{Simonovsky2017a}. A typical graph-based network is shown in Fig. \ref{fig:Graph_based_Network}. 

\textbf{Graph-based Methods in Spatial Domain.}
These methods define operations (e.g., convolution and pooling) in spatial domain. Specifically, convolution is usually implemented through MLP over spatial neighbors, and pooling is adopted to produce a new coarsened graph by aggregating information from each point's neighbors. Features at each vertex are usually assigned with coordinates, laser intensities or colors, while features at each edge are usually assigned with geometric attributes between two connected points.

As a pioneering work, Simonovsky et al. \cite{Simonovsky2017a} considered each point as a vertex of the graph, and connected each vertex to all its neighbors by a directed edge. Then, Edge-Conditioned Convolution (ECC) is proposed using a filter-generating network (e.g., MLP). Max pooling is adopted to aggregate neighborhood information and graph coarsening is implemented based on VoxelGrid \cite{rusu20113d}. In DGCNN \cite{Wang2019}, a graph is constructed in the feature space and dynamically updated after each layer of the network. As the core layer of EdgeConv, an MLP is used as the feature learning function for each edge, and channel-wise symmetric aggregation is applied onto the edge features associated with the neighbors of each point. Further, LDGCNN \cite{Zhang2019} removes the transformation network and links the hierarchical features from different layers in DGCNN \cite{Wang2019} to improve its performance and reduce the model size. An end-to-end unsupervised deep AutoEncoder network (namely, FoldingNet \cite{Yang2018}) is also proposed to use the concatenation of a vectorized local covariance matrix and point coordinates as its input. Inspired by Inception \cite{szegedy2015going} and DGCNN \cite{Wang2019}, Hassani and Haley \cite{Hassani2019} proposed an unsupervised multi-task autoencoder to learn point and shape features. The encoder is constructed based on mutli-scale graphs. The decoder is constructed using three unsupervised tasks including clustering, self-supervised classification and reconstruction, which are trained jointly with a mutli-task loss. Liu et al. \cite{Liu2019DPAM} proposed a Dynamic Points Agglomeration Module (DPAM) based on graph convolution to simplify the process of points agglomeration (sampling, grouping and pooling) into a simple step, which is implemented through multiplication of the agglomeration matrix and points feature matrix. Based on the PointNet architecture, a hierarchical learning architecture is constructed by stacking multiple DPAMs. Compared with the hierarchy strategy of PointNet++ \cite{PointNet++}, DPAM dynamically exploits the relation of points and agglomerates points in a semantic space. 

To exploit the local geometric structures, KCNet \cite{Shen2018} learns features based on kernel correlation. Specifically, a set of learnable points characterizing geometric types of local structures are defined as kernels. Then, affinity between the kernel and the neighborhood of a given point is calculated. In G3D \cite{Dominguez2018},  convolution is defined as a variant of polynomial of adjacency matrix, and pooling is defined as multiplying the Laplacian matrix and the vertex matrix by a coarsening matrix.
ClusterNet \cite{ClusterNet2019} utilizes a rigorously rotation-invariant module to extract rotation-invariant features from $k$ nearest neighbors for each point, and constructs hierarchical structures of a point cloud based on the unsupervised agglomerative hierarchical clustering method  with ward-linkage criteria \cite{mullner2011modern}. The features in each sub-cluster are first learned through an EdgeConv block and then aggregated through max pooling.

\why{To address the time-consuming problem of current data structuring methods (such as FPS and neighbor points querying), Xu et al. \cite{Xu_2020_CVPR} proposed to blend the advantages of volumetric based and point based methods to improve the computational efficiency. Experiments on the ModelNet classification task demonstrate that  the computational efficiency of the proposed Grid-GCN network is 5$\times$ faster than other models in average.}

\textbf{Graph-based Methods in Spectral Domain.}
These methods define convolutions as spectral filtering, which is implemented as the multiplication of signals on graph with eigenvectors of the graph Laplacian matrix \cite{Bruna2014, Defferrard2016}. 

RGCNN \cite{Rgcnn} constructs a graph by connecting each point with all other points in the point cloud and updates the graph Laplacian matrix in each layer. To make features of adjacent vertices more similar, a graph-signal smoothness prior is added into the loss function. To address the challenges caused by diverse graph topology of data, the SGC-LL layer in AGCN \cite{Li2018b} utilizes a learnable distance metric to parameterize the similarity between two vertices on the graph. The adjacency matrix obtained from graph is normalized using Gaussian kernels and learned distances. HGNN \cite{Feng2019HGNN} builds a hyperedge convolutional layer by applying spectral convolution on a hypergraph. 

Aforementioned methods operate on full graphs. To exploit local structural information, Wang et al. \cite{Wang2018a} proposed an end-to-end spectral convolution network LocalSpecGCN to work on a local graph (which is constructed from the $k$ nearest neighbors). This method does not require any offline computation of the graph Laplacian matrix and graph coarsening hierarchy. In PointGCN \cite{Zhang2018}, a graph is constructed based on $k$ nearest neighbors from a point cloud and each edge is weighted using a Gaussian kernel. Convolutional filters are defined as Chebyshev polynomials in  graph spectral domain. Global pooling and multi-resolution pooling are used to capture global and local features of the point cloud.  Pan et al. \cite{Pan2018} proposed 3DTI-Net by applying convolution on the $k$ nearest neighboring graphs in spectral domain. The invariance to geometry transformation is achieved by learning from relative Euclidean and direction distances.

\subsubsection{\bf \why{Hierarchical Data Structure-based Methods}}
These networks are constructed based on different \why{hierarchical data structures} (e.g., octree and kd-tree). In these methods, point features are learned hierarchically from leaf nodes to the root node along a tree.  

Lei et al. \cite{lei2019octree} proposed an octree guided CNN using spherical convolutional kernels (as described in Section \ref{3DGN}). Each layer of the network corresponds to one layer of the octree and a spherical convolutional kernel is applied at each layer. The values of neurons in the current layer are determined as the mean values of all relevant children nodes in the previous layer. Unlike OctNet \cite{Riegler2017} which is based on octree, Kd-Net \cite{Klokov2017} is built using multiple K-d trees with different splitting directions at each iteration. Following a bottom-up approach, the representation of a non-leaf node is computed from representations of its children using MLP. The feature of the root node (which describes the whole point cloud) is finally fed to fully connected layers to predict classification scores. Note that, Kd-Net shares parameters at each level according to the splitting type of nodes. 
3DContextNet \cite{Zeng2017} uses a standard balanced K-d tree to achieve feature learning and aggregation. At each level, point features are first learned through MLP based on local cues (which models inter-dependencies between points in a local region) and global contextual cues (which models the relationship for one position with respect to all other positions). Then, the feature of a non-leaf node is computed from its child nodes using MLP and aggregated by max pooling. For classification, the above process is repeated until the root node is attained. 

The hierarchy of SO-Net network is constructed by performing point-to-node $k$ nearest neighbor search \cite{Li2018a}. Specifically, a modified permutation invariant Self-Organizing Map (SOM) is used to model the spatial distribution of a point cloud. Individual point features are learned from normalized point-to-node coordinates through a series of fully connected layers. The feature of each node in SOM is extracted from point features associated with this node using channel-wise max pooling. The final feature is then learned from node features using an approach similar to PointNet \cite{PointNet}. Compared to PointNet++ \cite{PointNet++}, the hierarchy of SOM is more efficient and the spatial distribution of the point cloud is fully explored. 

\begin{table*}[]
\centering
\caption{Comparative 3D shape classification results on the ModelNet10/40 benchmarks. Here, we only focus on point-based networks. `\#params' represents the number of parameters of a model, `OA' represents \why{the mean accuracy for all test instances} and `mAcc' represents the mean accuracy \why{ for all shape classes} in the table. The symbol `-' means the results are unavailable.}
\label{Tab:ModelNet}
\resizebox{\textwidth}{!}{%
\begin{tabular}{|c|r|c|c|c|c|c|c|}
\hline
\multicolumn{2}{|c|}{\bf Methods} & \bf Input & \bf \#params (M) & \begin{tabular}[c]{@{}c@{}}\bf ModelNet40\\ \bf (OA)\end{tabular} & \begin{tabular}[c]{@{}c@{}}\bf ModelNet40\\ \bf (mAcc)\end{tabular} & \begin{tabular}[c]{@{}c@{}}\bf ModelNet10\\ \bf (OA)\end{tabular} & \begin{tabular}[c]{@{}c@{}}\bf ModelNet10\\\bf (mAcc)\end{tabular} \\ \hline
\multirow{8}{*}{\begin{tabular}[c]{@{}c@{}}Pointwise MLP \\ Methods\end{tabular}} 
& PointNet \cite{PointNet} & Coordinates & 3.48 & 89.2\% & 86.2\% & - & - \\ \cline{2-8} 
& PointNet++ \cite{PointNet++} & Coordinates & 1.48 & 90.7\% & - & - & - \\ \cline{2-8}
& MO-Net \cite{Joseph-Rivlin2018} & Coordinates & 3.1 & 89.3\% & 86.1\% & - & - \\ \cline{2-8}
& Deep Sets \cite{Zaheer2017} & Coordinates & - & 87.1\% & - & - & - \\ \cline{2-8}
& PAT \cite{Yang2019} & Coordinates & - & 91.7\% & - & - & -\\ \cline{2-8}
& PointWeb \cite{PointWeb2019} & Coordinates & - & 92.3\% & 89.4\% & - & -\\ \cline{2-8}
& SRN-PointNet++  \cite{Duan2019} & Coordinates & - & 91.5\% & - & - & -\\ \cline{2-8}
& JUSTLOOKUP \cite{Lin2019Justlookup} & Coordinates & - & 89.5\% & 86.4\% & 92.9\% & 92.1\%\\ \cline{2-8}
& \why{PointASNL \cite{Yan_2020_CVPR}}  & Coordinates & - & 92.9\% & - & 95.7\% & -\\ \cline{2-8}
& \why{PointASNL \cite{Yan_2020_CVPR}}  & Coordinates+Normals & - & 93.2\% & - & 95.9\% & -\\ \cline{2-8}
\hline
\multirow{20}{*}{\begin{tabular}[c]{@{}c@{}}Convolution-based\\ Methods\end{tabular}} 
& Pointwise-CNN \cite{Hua2018} & Coordinates & - & 86.1\% & 81.4\% & - & -\\ \cline{2-8}
& PointConv \cite{Wu2018} & Coordinates+Normals & - & 92.5\% & - & - & -\\ \cline{2-8}
& MC Convolution \cite{Hermosilla2018} & Coordinates & - & 90.9\% & - & - & -\\ \cline{2-8}
& SpiderCNN \cite{Xu2018} & Coordinates+Normals & - & 92.4\% & - & - & -\\ \cline{2-8}
& PointCNN \cite{li2018pointcnn} & Coordinates & 0.45 & 92.2\% & 88.1\% & - & -\\ \cline{2-8}
& Flex-Convolution \cite{Flexconv} & Coordinates & - & 90.2\% & - & - & -\\ \cline{2-8}
& PCNN \cite{Matan2018} & Coordinates & 1.4 & 92.3\% & - & 94.9\% & -\\ \cline{2-8}
& Boulch \cite{Boulch2019} & Coordinates & - & 91.6\% & 88.1\% & - & -\\ \cline{2-8}
& RS-CNN \cite{Liu2019} & Coordinates & - & 93.6\% & - & - & -\\ \cline{2-8}
& Spherical CNNs \cite{Esteves2017} & Coordinates & 0.5 & 88.9\% & - & - & -\\ \cline{2-8}
& GeoCNN \cite{Lan2018} & Coordinates & - & 93.4\% & 91.1\% & - & -\\ \cline{2-8}
& $\Psi$-CNN \cite{lei2019octree} & Coordinates & - & 92.0\% & 88.7\% & 94.6\% & 94.4\%\\ \cline{2-8}
& A-CNN \cite{Komarichev2019} & Coordinates & - & 92.6\% & 90.3\% & 95.5\% & 95.3\%\\ \cline{2-8}
& SFCNN \cite{SFCNN2019} & Coordinates & - & 91.4\% & - & - & -\\ \cline{2-8}
& SFCNN \cite{SFCNN2019} & Coordinates+Normals & - & 92.3\% & - & - & -\\ \cline{2-8}
& DensePoint \cite{Densepoint2019} & Coordinates  & 0.53 & 93.2\% & - & 96.6\% & -\\ \cline{2-8}
& KPConv rigid \cite{kpconv} & Coordinates & - & 92.9\% & - & - & -\\ \cline{2-8}
& KPConv deform \cite{kpconv} & Coordinates & - & 92.7\% & - & - & -\\ \cline{2-8}
& InterpCNN \cite{Mao2019InterpConv} & Coordinates & 12.8 & 93.0\% & - & - & -\\ \cline{2-8}
& ConvPoint \cite{Boulch2019ConvPoint} & Coordinates & - & 91.8\% & 88.5\% & - & -\\ \hline 
\multirow{11}{*}{\begin{tabular}[c]{@{}c@{}}Graph-based\\ Methods\end{tabular}} 
& ECC \cite{Simonovsky2017a} & Coordinates & - & 87.4\% & 83.2\% & 90.8\% & 90.0\%\\ \cline{2-8}
& KCNet \cite{Shen2018} & Coordinates & 0.9 & 91.0\% & - & 94.4\% & -\\ \cline{2-8}
& DGCNN \cite{Wang2019} & Coordinates & 1.84 & 92.2\% & 90.2\% & - & -\\ \cline{2-8}
& LocalSpecGCN \cite{Wang2018a} & Coordinates+Normals & - & 92.1\% & - & - & -\\ \cline{2-8}
& RGCNN \cite{Rgcnn} & Coordinates+Normals & 2.24 & 90.5\% & 87.3\% & - & -\\ \cline{2-8}
& LDGCNN \cite{Zhang2019} & Coordinates & - & 92.9\% & 90.3\% & - & -\\ \cline{2-8}
& 3DTI-Net \cite{Pan2018} & Coordinates & 2.6 & 91.7\% & - & - & -\\ \cline{2-8}
& PointGCN \cite{Zhang2018} & Coordinates & - & 89.5\% & 86.1\% & 91.9\% & 91.6\%\\ \cline{2-8}
& ClusterNet \cite{ClusterNet2019} & Coordinates & - & 87.1\% & - & - & -\\ \cline{2-8}
& Hassani et al. \cite{Hassani2019} & Coordinates & - & 89.1\% & - & - & -\\ \cline{2-8}
& DPAM \cite{Liu2019DPAM} & Coordinates & - & 91.9\% & 89.9\% & 94.6\% & 94.3\%\\ \cline{2-8}
& \why{Grid-GCN \cite{Xu_2020_CVPR}} & Coordinates & - & 93.1\% & 91.3\% & 97.5\% & 97.4\%\\ \cline{2-8}
\hline 
\multirow{6}{*}{\begin{tabular}[c]{@{}c@{}}\why{Hierarchical Data Structure} \\-based Methods\end{tabular}} 
& KD-Net \cite{Klokov2017} & Coordinates & 2.0  & 91.8\% & 88.5\% & 94.0\% & 93.5\%\\ \cline{2-8}
& SO-Net \cite{Li2018a} & Coordinates & - & 90.9\% & 87.3\% & 94.1\% & 93.9\%\\ \cline{2-8}
& SCN \cite{Xie2018a} & Coordinates & - & 90.0\% & 87.6\% & - & -\\ \cline{2-8}
& A-SCN \cite{Xie2018a} & Coordinates & - & 89.8\% & 87.4\% & - & -\\ \cline{2-8}
& 3DContextNet \cite{Zeng2017} & Coordinates & - & 90.2\% & - & - & -\\ \cline{2-8}
& 3DContextNet \cite{Zeng2017} & Coordinates+Normals & - & 91.1\% & - & - & -\\ \hline 
\multirow{9}{*}{Other Methods}
& 3DmFV-Net \cite{Ben-Shabat2017} & Coordinates & 4.6 & 91.6\% & - & 95.2\% & -\\ \cline{2-8}
& PVNet \cite{You2018} & Coordinates+Views & - & 93.2\% & - & - & -\\ \cline{2-8}
& PVRNet \cite{You2018a}  & Coordinates+Views & - & 93.6\% & - & - & -\\ \cline{2-8}
& 3DPointCapsNet \cite{Zhao2019} & Coordinates & - & 89.3\% & - & - & -\\ \cline{2-8}
& DeepRBFNet \cite{Chen2018} & Coordinates & 3.2 & 90.2\% & 87.8\% & - & -\\ \cline{2-8}
& DeepRBFNet \cite{Chen2018} & Coordinates+Normals & 3.2 & 92.1\% & 88.8\% & - & -\\ \cline{2-8}
& Point2Sequences \cite{Liu2019Point2Sequencces} & Coordinates & - & 92.6\% & 90.4\% & 95.3\% & 95.1\%\\ \cline{2-8}
& RCNet \cite{Wu2019RCNet} & Coordinates & - & 91.6\% & - & 94.7\% & -\\ \cline{2-8}
& RCNet-E \cite{Wu2019RCNet} & Coordinates & - & 92.3\% & - & 95.6\% & -\\ \hline 
\end{tabular}%
}
\end{table*}

\subsubsection{\bf Other Methods}
In addition, many other schemes have also been proposed. 
RBFNet \cite{Chen2018} explicitly models the spatial distribution of points by aggregating features from sparsely distributed Radial Basis Function (RBF) kernels with learnable kernel positions and sizes. 3DPointCapsNet \cite{Zhao2019} learns point independent features with pointwise MLP and convolutional layers, and extracts global latent representation with multiple max-pooling layers. Based on unsupervised dynamic routing, powerful representative latent capsules are then learned. \why{Qin et al. \cite{Qin2019PointDAN} proposed an end-to-end unsupervised domain adaptation network PointDAN for 3D point cloud representation. To capture semantic properties of a point cloud, a self-supervised method is proposed to reconstruct the point cloud, whose parts have been randomly rearranged  \cite{Sievers2019}. Li et al. \cite{Li_2020_CVPR} proposed an auto-augmentation framework, PointAugment, to automatically optimize and augment point cloud samples for network training. Specifically, shape-wise transformation and point-wise displacement for each input sample are automatically learned, and the network is trained by alternatively optimizing and updating the learnable parameters of its augmentor and classifier. Inspired by shape context \cite{belongie2002shape}, Xie et al. \cite{Xie2018a} proposed a ShapeContextNet architecture by combining affinity point selection and compact feature aggregation into a soft alignment operation using dot-product self-attention \cite{all_you_need}. To handle noise and occlusion in 3D point clouds, Bobkov et al. \cite{Bobkov2018}  fed handcrafted point pair function based 4D rotation invariant descriptors into a 4D convolutional neural network. Prokudin et al. \cite{Prokudin2019} first randomly sampled a basis point set with a uniform distribution from a unit ball, and then encoded a point cloud as minimal distances to the basis point set. Consequently, the point cloud is converted to a vector with a relatively small fixed length. The encoded representation can then be processed with existing machine learning methods.}

RCNet \cite{Wu2019RCNet} utilizes standard RNN and 2D CNN to construct a permutation-invariant network for 3D point cloud processing. The point cloud is first partitioned into parallel beams and sorted along a specific dimension, and each beam is then fed into a shared RNN. The learned features are further fed into an efficient 2D CNN for hierarchical feature aggregation. To enhance its description ability, RCNet-E is proposed to ensemble multiple RCNets along different partition and sorting directions. Point2Sequences \cite{Liu2019Point2Sequencces} is another RNN-based model that captures correlations between different areas in local regions of point clouds. It considers features learned from a local region at multiple scales as sequences and feeds these sequences from all local regions into an RNN-based encoder-decoder structure to aggregate local region features. 

Several methods also learn from both 3D point clouds and 2D images. In PVNet \cite{You2018}, high-level global features extracted from multi-view images are projected into the subspace of point clouds through an embedding network, and fused with point cloud features through a soft attention mask. Finally, a residual connection is employed for fused features and multi-view features to perform shape recognition. Later, PVRNet \cite{You2018a} is further proposed to exploit the relation between a 3D point cloud and its multiple views by a relation score module. Based on the relation scores, the original 2D global view features are enhanced for point-single-view fusion and point-multi-view fusion. 

\subsection{Summary}
The ModelNet10/40 \cite{wu20153d} datasets are the most frequently used datasets for 3D shape classification. Table \ref{Tab:ModelNet} shows the results achieved by different point-based networks. Several observations can be drawn:
\begin{itemize}
\item[$\bullet$] Pointwise MLP networks are usually served as the basic building block for other types of networks to learn pointwise features. 

\item[$\bullet$] As a standard deep learning architecture, convolution-based networks can achieve superior performance on  irregular 3D point clouds. More attention should be paid to both discrete and continuous convolution networks for irregular data.

\item[$\bullet$] Due to its inherent strong capability to handle irregular data, graph-based networks have attracted increasingly more attention in recent years. However, it is still challenging to extend graph-based networks in the spectral domain to various graph structures. 

\end{itemize}

\section{3D Object Detection and Tracking} \label{sec:object_detection}
In this section, we will review existing methods for 3D object detection, 3D object tracking and 3D scene flow estimation.

\subsection{3D Object Detection}
\begin{figure}[h]
\noindent \begin{centering}
\noindent\subfloat[{ScanNetV2 \cite{ScanNet} dataset}]{\noindent \begin{centering}
\includegraphics[scale=1]{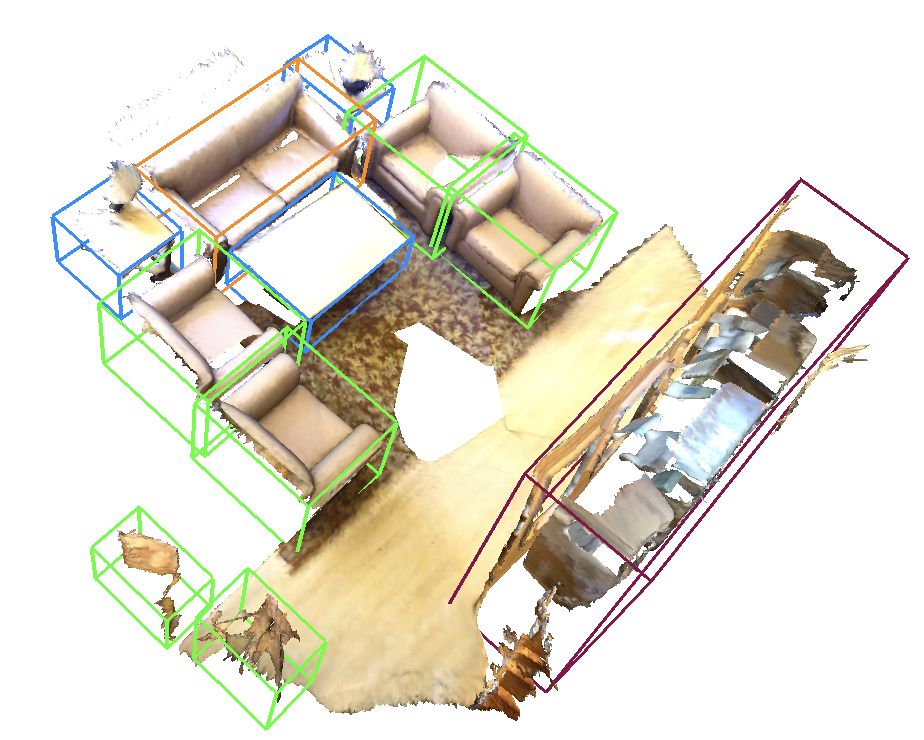}
\par\end{centering}
}
\subfloat[{KITTI \cite{KITTI} dataset}]{\noindent \begin{centering}
\includegraphics[scale=0.2]{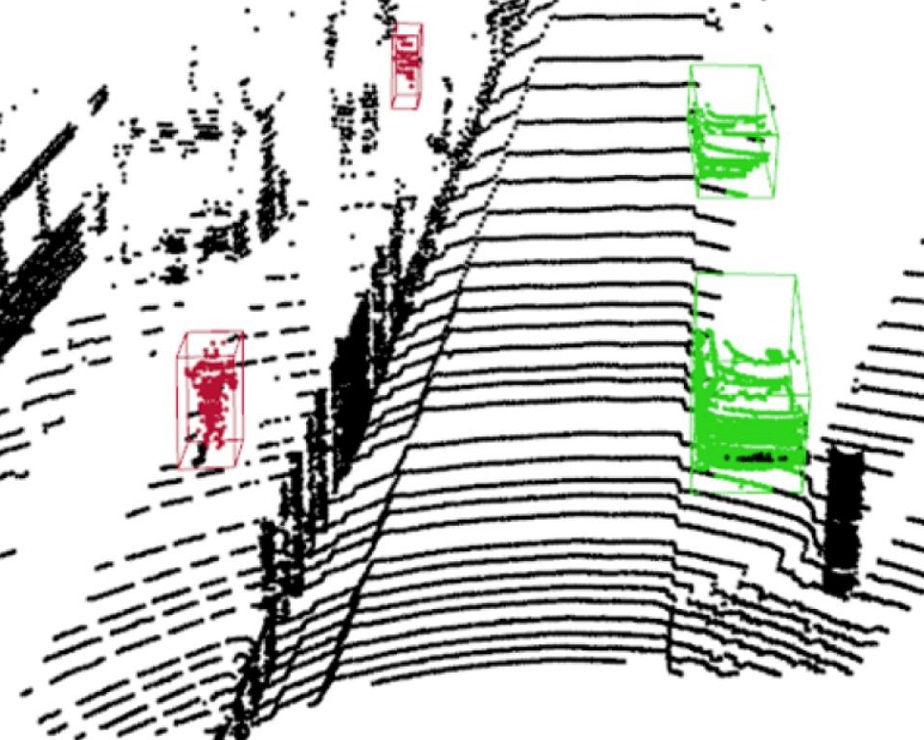}
\par\end{centering}
}
\par\end{centering}
\caption{\label{fig:object-detection} An illustration of 3D object detection. (a) and (b) are originally shown in \cite{VoteNet} and \cite{PointGNN}, respectively.}
\end{figure}

\begin{figure*}[t]
\centering
\includegraphics[width=6.5in]{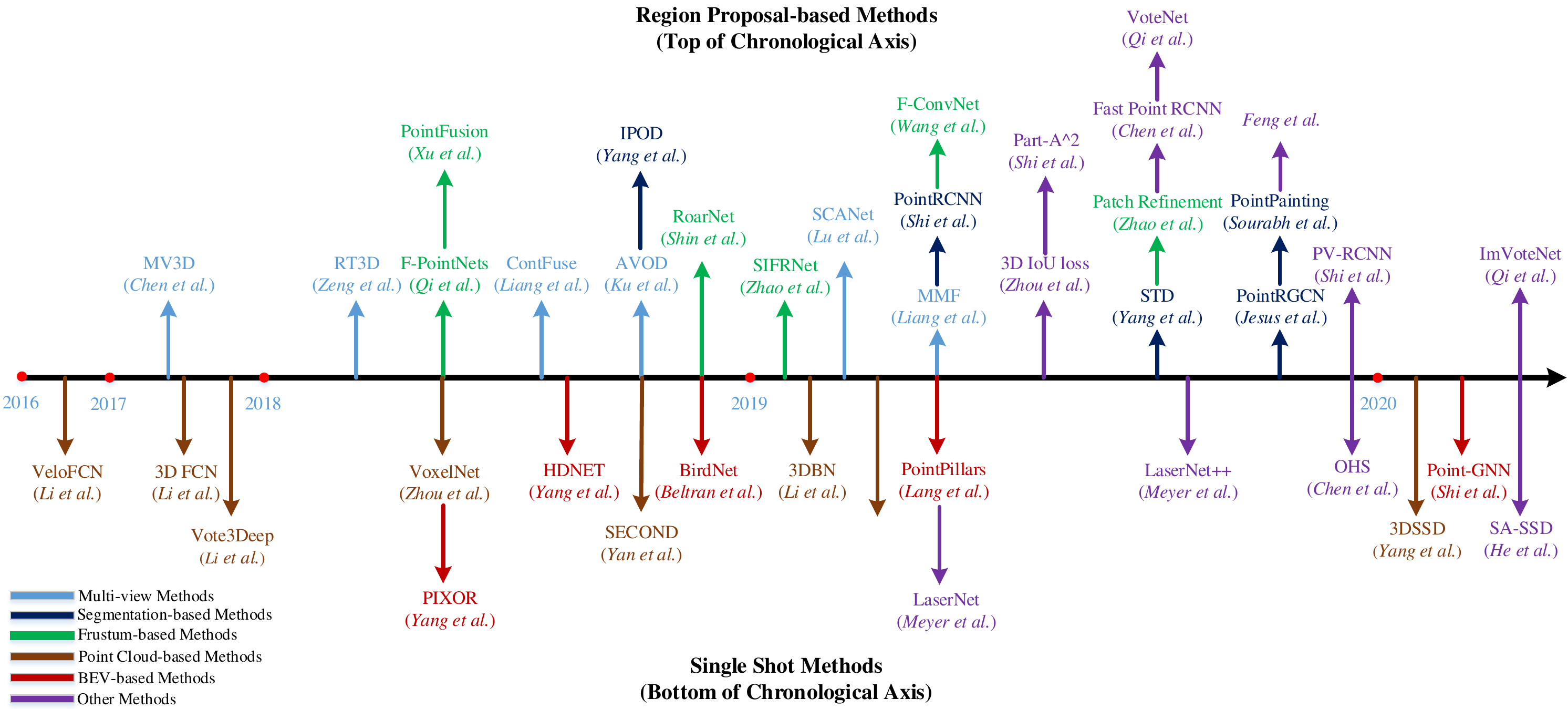}
\caption{Chronological overview of the most relevant deep learning-based 3D object detection methods.\label{fig:milestone-detection}}
\end{figure*}

\hao{A typical 3D object detector takes the point cloud of a scene as its input and produces an oriented 3D bounding box around each detected object, as shown in Fig. \ref{fig:object-detection}.} Similar to object detection in images \cite{liu2018deep}, 3D object detection methods can be divided into two categories: region proposal-based and single shot methods. Several milestone methods are presented in Fig. \ref{fig:milestone-detection}.

\subsubsection{\bf Region Proposal-based Methods}
These methods first propose several possible regions (also called proposals) containing objects, and then extract region-wise features to determine the category label of each proposal. According to their object proposal generation approach, these methods can further be divided into three categories: multi-view based, segmentation-based and frustum-based methods.

\begin{figure}[h]
\centering
\includegraphics[width=\columnwidth,keepaspectratio]{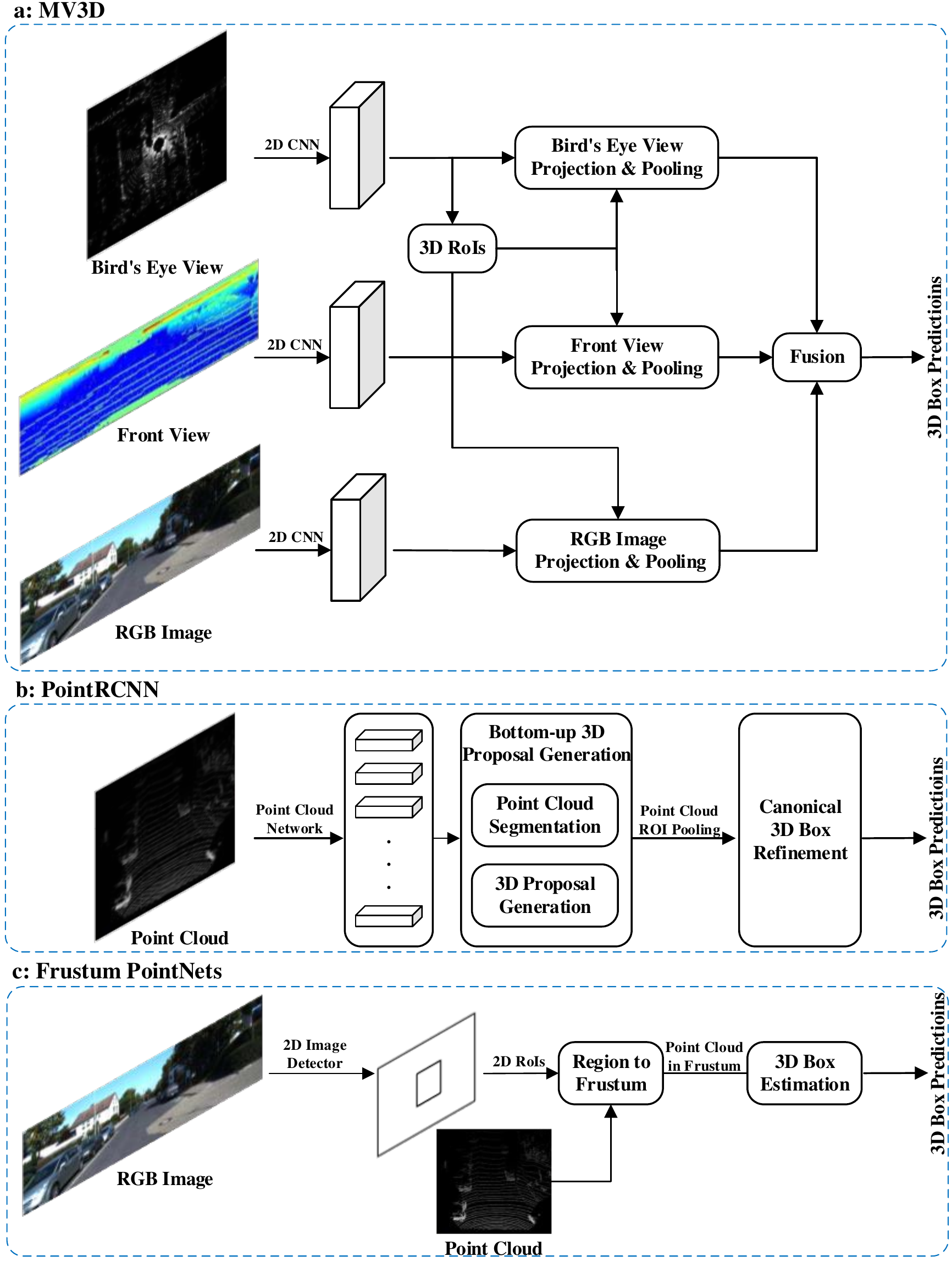}
\caption{Typical networks for three categories of region proposal-based 3D object detection methods. From top to bottom: (a) multi-view based, (b) segmentation-based and (c) frustum-based methods.}
\label{fig:detection_frameworks}
\end{figure}

\textbf{Multi-view based Methods.} These methods fuse proposal-wise features from different view maps (e.g., LiDAR front view, Bird's Eye View (BEV), and image) to obtain 3D rotated boxes, as shown in Fig. \ref{fig:detection_frameworks}(a). The computational cost of these methods is usually high. 

Chen et al. \cite{chen2017multi} generated a group of highly accurate 3D candidate boxes from the BEV map and projected them to the feature maps of multiple views (e.g., LiDAR front view image, RGB image). They then combined these region-wise features from different views to predict oriented 3D bounding boxes, as shown in Fig. \ref{fig:detection_frameworks}(a). Although this method achieves a recall of 99.1\% at an Intersection over Union (IoU) of 0.25 with only 300 proposals, its speed is too slow for practical applications. Subsequently, several approaches have been developed to improve multi-view 3D object detection methods from two aspects. 

\textbf{First}, several methods have been proposed to efficiently fuse the information of different modalities. To generate 3D proposals with a high recall for small objects, Ku et al. \cite{AVOD} proposed a multi-modal fusion-based region proposal network. They first extracted equal-sized features from both BEV and image views using cropping and resizing operations, and then fused these features using element-wise mean pooling. Liang et al. \cite{ContFuse} exploited continuous convolutions to enable effective fusion of image and 3D LiDAR feature maps at different resolutions. Specifically, they extracted nearest corresponding image features for each point in the BEV space and then used bilinear interpolation to obtain a dense BEV feature map by projecting image features into the BEV plane. Experimental results show that dense BEV feature maps are more suitable for 3D object detection than discrete image feature maps and sparse LiDAR feature maps. Liang et al. \cite{MMF} presented a multi-task multi-sensor 3D object detection network for end-to-end training. Specifically, multiple tasks (e.g., 2D object detection, ground estimation and depth completion) are exploited to help the network learn better feature representations. 
The learned cross-modality representation is further exploited to produce highly accurate object detection results. Experimental results show that this method achieves a significant improvement on 2D, 3D and BEV detection tasks, and outperforms previous state-of-the-art methods on the TOR4D  benchmark \cite{PIXOR,FastandFurious}.

\textbf{Second}, different methods have been investigated to extract robust representations of the input data. Lu et al. \cite{Scanet} explored multi-scale contextual information by introducing a Spatial Channel Attention (SCA) module, which captures the global and multi-scale context of a scene and highlights useful features. They also proposed an Extension Spatial Unsample (ESU) module to obtain high-level features with rich spatial information by combining multi-scale low-level features, thus generating reliable 3D object proposals. Although better detection performance can be achieved, the aforementioned multi-view methods take a long runtime since they perform feature pooling for each proposal. Subsequently, Zeng et al. \cite{RT3D} used a pre-RoI pooling convolution to improve the  efficiency of \cite{chen2017multi}. Specifically, they moved the majority of convolution operations to be ahead of the RoI pooling module. Therefore, RoI convolutions are performed once for all object proposals. 
Experimental results show that this method can run at a speed of 11.1 fps, which is 5 times faster than MV3D \cite{chen2017multi}.

\textbf{Segmentation-based Methods.} These methods first leverage existing semantic segmentation techniques to remove most background points, and then generate a large amount of high-quality proposals on foreground points to save computation, as shown in Fig. \ref{fig:detection_frameworks}(b). Compared to multi-view methods \cite{chen2017multi,AVOD,RT3D}, these methods achieve higher object recall rates and are more suitable for complicated scenes with highly occluded and crowded objects.

Yang et al. \cite{IPOD} used a 2D segmentation network to predict foreground pixels and projected them into point clouds to remove most background points. They then generated proposals on the predicted foreground points and designed a new criterion named PointsIoU to reduce the redundancy and ambiguity of proposals. Following \cite{IPOD}, Shi et al. \cite{PointRCNN} proposed a PointRCNN framework. Specifically, they directly segmented 3D point clouds to obtain foreground points and then fused semantic features and local spatial features to produce high-quality 3D boxes. Following the Region Proposal Network (RPN) stage of \cite{PointRCNN}, Jesus et al. \cite{PointRGCN} proposed a pioneering work to leverage Graph Convolution Network (GCN) for 3D object detection. Specifically, two modules are introduced to refine object proposals using graph convolution. The first module R-GCN utilizes all points contained in a proposal to achieve per-proposal feature aggregation. The second module C-GCN fuses per-frame information from all proposals to regress accurate object boxes by exploiting contexts. 
Sourabh et al. \cite{PointPainting} projected a point cloud into the output of the image-based segmentation network and appended the semantic prediction scores to the points. The painted points are fed into existing detectors \cite{PointRCNN,VoxelNet,PointPillars} to achieve significant performance improvement. Yang et al. \cite{STD} associated each point with a spherical anchor. 
The semantic score of each point is then used to remove redundant anchors. Consequently, this method achieves a higher recall with lower computational cost as compared to previous methods \cite{IPOD,PointRCNN}. In addition, a PointsPool layer is proposed to learn compact features for interior points in proposals and a parallel IoU branch is introduced to improve localization accuracy and detection performance. 

\textbf{Frustum-based Methods.} These methods first leverage existing 2D object detectors to generate 2D candidate regions of objects and then extract a 3D frustum proposal for each 2D candidate region, as shown in Fig. \ref{fig:detection_frameworks}(c). Although these methods can efficiently propose possible locations of 3D objects, the step-by-step pipeline makes their performance limited by 2D image detectors.

F-PointNets \cite{F-PointNet} is a pioneering work in this direction. It generates a frustum proposal for each 2D region and applies PointNet \cite{PointNet} (or PointNet++ \cite{PointNet++}) to learn point cloud features of each 3D frustum for amodal 3D box estimation. 
In a follow-up work, Zhao et al. \cite{SIFRNet} proposed a Point-SENet module to predict a set of scaling factors, which were further used to adaptively highlight useful features and suppress informative-less features. They also integrated the PointSIFT \cite{PointSIFT} module into the network to capture orientation information of point clouds, which achieved strong robustness to shape scaling. This method achieves significant improvement on both indoor and outdoor datasets \cite{KITTI,SunRGBD} as compared to F-PointNets \cite{F-PointNet}. 

Xu et al. \cite{PointFusion} leveraged both 2D image region and its corresponding frustum points to accurately regress 3D boxes. 
To fuse image features and global features of point clouds, they presented a global fusion network for direct regression of box corner locations. They also proposed a dense fusion network for the prediction of point-wise offsets to each corner. 
Shin et al. \cite{Roarnet} first estimated 2D bounding boxes and 3D poses of objects from a 2D image, and then extracted multiple geometrically feasible object candidates. These 3D candidates are fed into a box regression network to predict accurate 3D object boxes. Wang et al. \cite{F-ConvNet} generated a sequence of frustums along the frustum axis for each 2D region and applied PointNet \cite{PointNet} to extract features for each frustum. The frustum-level features are reformed to generate a 2D feature map, which is then fed into a fully convolutional network for 3D box estimation. This method achieves the state-of-the-art performance among 2D image-based methods and was ranked in the top position of the official KITTI leaderboard. \why{Johannes} et al. \cite{Patch} first obtained a preliminary detection results on the BEV map, and then extracted small point subsets (also called patches) based on the BEV predictions. A local refinement network is applied to learn the local features of patches to predict highly accurate 3D bounding boxes.

\begin{table*}[t]
\centering
\caption{Comparative 3D object detection results on the KITTI test 3D detection benchmark. 3D bounding box IoU threshold is 0.7 for cars and 0.5 for pedestrians and cyclists. The modalities are LiDAR (L) and image (I). `E', `M' and `H' represent easy, moderate and hard classes of objects, respectively. For simplicity, we omit the `\%' after the value. The symbol `-' means the results are unavailable.}
\label{Tab:KITTI3D}
\resizebox{\textwidth}{!}{
\begin{tabular}{|c|c|r|c|c|ccc|ccc|ccc|}
\hline
 \multicolumn{3}{|c|}{\multirow{2}{*}{\bf Method}} & \multirow{2}{*}{\bf Modality} & \multirow{2}{*}{\bf \tabincell{c}{Speed \\ (fps)}} & \multicolumn{3}{|c|}{\bf Cars} & \multicolumn{3}{|c|}{\bf Pedestrians} & \multicolumn{3}{|c|}{\bf Cyclists} \\\cline{6-14}
 \multicolumn{3}{|c|}{} & & & {\bf E} & {\bf M} & {\bf H} & {\bf E} & {\bf M} & {\bf H} & {\bf E} & {\bf M} & {\bf H} \\\hline
\multirow{19}{*}{\tabincell{c}{Region \\ Proposal \\-based \\ Methods}} & \multirow{6}{*}{\tabincell{c}{Multi-view \\ Methods}} & {MV3D \cite{chen2017multi}} & {L \& I} & 2.8 & 74.97 & 63.63 & 54.00 & - & - & - & - & - & - \\
  & & {AVOD \cite{AVOD}} & {L \& I} & 12.5 & 76.39 & 66.47 & 60.23 & 36.10 & 27.86 & 25.76 & 57.19 & 42.08 & 38.29 \\
  & & {ContFuse \cite{ContFuse}} & {L \& I} & 16.7 & 83.68 & 68.78 & 61.67 & - & - & - & - & - & - \\
  & & {MMF \cite{MMF}} & {L \& I} & 12.5 & 88.40 & 77.43 & 70.22 & - & - & - & - & - & - \\
  & & {SCANet \cite{Scanet}} & {L \& I} & 11.1 & 79.22 & 67.13 & 60.65 & - & - & - & - & - & - \\
  & & {RT3D \cite{RT3D}} & {L \& I} & 11.1 & 23.74 & 19.14 & 18.86 & - & - & - & - & - & - \\\cline{2-14}
  & \multirow{3}{*}{\tabincell{c}{Segmentation \\ -based \\ Methods}} & {IPOD \cite{IPOD}} & {L \& I} & 5.0 & 80.30 & 73.04 & 68.73 & 55.07 & 44.37 & 40.05 & 71.99 & 52.23 & 46.50 \\
  & & {PointRCNN \cite{PointRCNN}} & {L} & 10.0 & 86.96 & 75.64 & 70.70 & 47.98 & 39.37 & 36.01 & 74.96 & 58.82 & 52.53 \\
  & & {PointRGCN \cite{PointRGCN}} & {L} & 3.8 & 85.97 & 75.73 & 70.60 & - & - & - & - & - & - \\
  & & {PointPainting \cite{PointPainting}} & {L \& I} & 2.5 & 82.11 & 71.70 & 	67.08 & 50.32 & 40.97 & 37.87 & 77.63 & 63.78 & 55.89 \\
  & & {STD \cite{STD}} & {L} & 12.5 & 87.95 & 79.71 & 75.09 & 53.29 & 42.47 & 38.35 & 78.69 & 61.59 & 55.30 \\\cline{2-14}
  & \multirow{6}{*}{\tabincell{c}{Frustum \\ -based \\ Methods}} & {F-PointNets \cite{F-PointNet}} & {L \& I} & 5.9 & 82.19 & 69.79 & 60.59 & 50.53 & 42.15 & 38.08 & 72.27 & 56.12 & 49.01 \\
  & & {SIFRNet \cite{SIFRNet}} & {L \& I} & - & - & - & - & - & - & - & - & - & - \\
  & & {PointFusion \cite{PointFusion}} & {L \& I} & - & 77.92 &   63.00 & 53.27 & 33.36 & 28.04 & 23.38 & 49.34 & 29.42 & 26.98 \\
  & & {RoarNet \cite{Roarnet}} & {L \& I} & 10.0 & 83.71 & 73.04 & 59.16 & - & - & - & - & - & - \\
  & & {F-ConvNet \cite{F-ConvNet}} & {L \& I} & 2.1 & 87.36 & 76.39 & 66.69 & 52.16 & 43.38 & 38.80 & 81.98 & 65.07 & 56.54 \\
  & & {\tabincell{c}{Patch Refinement \cite{Patch}}} & {L} & 6.7 & 88.67 & 77.20 & 71.82 & - & - & - & - & - & - \\\cline{2-14}
  & \multirow{4}{*}{\tabincell{c}{Other \\ Methods}} & {3D IoU loss \cite{3D-IoU}} & {L} & 12.5 & 86.16 & 76.50 & 71.39 & - & - & - & - & - & - \\
  & & {\tabincell{c}{Fast Point R-CNN \cite{Fast-PointRCNN}}} & {L} & 16.7 & 84.80 & 74.59 & 67.27 & - & - & - & - & - & - \\
  & & {PV-RCNN \cite{PVRCNN}} & {L} & 12.5 & 90.25 & 81.43 & 76.82 & - & - & - & - & - & - \\
  & & {VoteNet \cite{VoteNet}} & {L} & - & - & - & - & - & - & - & - & - & - \\
  & & {Feng et al. \cite{VoteNet2}} & {L} & - & - & - & - & - & - & - & - & - & - \\
  & & {ImVoteNet \cite{ImVoteNet}} & {L \& I} & - & - & - & - & - & - & - & - & - & - \\
  & & {Part-A\^{}2 \cite{PartA2}} & {L} & 12.5 & 87.81 & 78.49 & 73.51 & - & - & - & - & - & - \\\hline
  \multirow{13}{*}{\tabincell{c}{Single \\ Shot \\ Methods}} & \multirow{3}{*}{\tabincell{c}{BEV-based \\ Methods}} & {PIXOR \cite{PIXOR}} & {L} & 28.6 & - & - & - & - & - & - & - & - & - \\
  & & {HDNET \cite{HDNET}} & {L} & 20.0 & - & - & - & - & - & - & - & - & - \\
  & & {BirdNet \cite{BirdNet}} & {L} & 9.1 & 13.53 & 9.47 & 8.49 & 12.25 & 8.99 & 8.06 & 16.63 & 10.46 & 9.53 \\\cline{2-14}
   & \multirow{8}{*}{\tabincell{c}{Discretization \\ -based \\ Methods}} &  {VeloFCN \cite{VeloFCN}} & {L} & 1.0 & - & - & - & - & - & - & - & - & - \\
  & & {3D FCN \cite{3DFCN}} & {L} & {\textless0.2} & - & - & - & - & - & - & - & - & - \\
  & & {Vote3Deep \cite{Vote3deep}} & {L} & - & - & - & - & - & - & - & - & - & - \\
  & & {3DBN \cite{3DBN}} & {L} & 7.7 & 83.77 & 73.53 & 66.23 & - & - & - & - & - & - \\
  & & {VoxelNet \cite{VoxelNet}} & {L} & 2.0 & 77.47 & 65.11 & 57.73 & 39.48 & 33.69 & 31.51 & 61.22 & 48.36 & 44.37 \\
  & & {SECOND \cite{SECOND}} & {L} & 26.3 & 83.34 & 72.55 & 65.82 & 48.96 & 38.78 & 34.91 & 71.33 & 52.08 & 45.83 \\
  & & {MVX-Net \cite{MVX-Net}} & {L \& I} & 16.7 & 84.99 & 71.95 & 64.88 & - & - & - & - & - & - \\
  & & {PointPillars \cite{PointPillars}} & {L} & 62.0 & 82.58 & 74.31 & 68.99 & 51.45 & 41.92 & 38.89 & 77.10 & 58.65 & 51.92 \\
  & & {SA-SSD \cite{SA-SSD}} & {L} & 25.0 & 88.75 & 79.79 & 74.16 & - & - & - & - & - & - \\\cline{2-14}
  & \tabincell{c}{Point-based \\ Methods} & {3DSSD \cite{3DSSD}} & {L} & 25.0 & 88.36 & 79.57 & 74.55 & 54.64 & 44.27 & 40.23 & 82.48 & 64.10 & 56.90 \\\hline
  & \multirow{2}{*}{\tabincell{c}{Other \\ Methods}} & {LaserNet \cite{LaserNet}} & {L} & 83.3 & - & - & - & - & - & - & - & - & - \\
  & & {LaserNet++ \cite{LaserNet++}} & {L \& I} & 26.3 & - & - & - & - & - & - & - & - & - \\
  & & {OHS-Dense \cite{OHS}} & {L} & 33.3 & 88.12 & 78.34 & 73.49 & 47.14 & 39.72 & 37.25 & 79.09 & 62.72 & 56.76 \\
  & & {OHS-Direct \cite{OHS}} & {L} & 33.3 & 86.40 & 77.74 & 72.97 & 51.29 & 44.81 & 41.13 & 77.70 & 63.16 & 57.16 \\
  & & {Point-GNN \cite{PointGNN}} & {L} & 1.7 & 88.33 & 79.47 & 72.29 & 51.92 & 43.77 & 40.14 & 78.60 & 63.48 & 57.08\\\hline
\end{tabular}}
\end{table*}

\textbf{Other Methods.} Motivated by the success of axis-aligned IoU in object detection in images, Zhou et al. \cite{3D-IoU} integrated the IoU of two 3D rotated bounding boxes into several state-of-the-art detectors \cite{SECOND,PointPillars,PointRCNN} to achieve consistent performance improvement. Chen et al. \cite{Fast-PointRCNN} proposed a two-stage network architecture to use both point cloud and voxel representations. First, point clouds are voxelized and fed to a 3D backbone network to produce initial detection results. Second, the interior point features of initial predictions are further exploited for box refinements. Although this design is conceptually simple, it achieves comparable performance to \cite{PointRCNN} while maintaining a speed of 16.7 fps. 
Shi et al. \cite{PVRCNN} proposed PointVoxel-RCNN (PV-RCNN) to leverage both 3D convolutional network and PointNet-based set abstraction for the learning of point cloud features. Specifically, the input point clouds are first voxelized and then fed into a 3D sparse convolutional network to generate high-quality proposals. The learned voxel-wise features are then encoded into a small set of key points via a voxel set abstraction module. In addition, they also proposed a keypoint-to-grid ROI abstraction module to capture rich context information for box refinement. Experimental results show that this method outperforms previous methods by a remarkable margin and is ranked first\footnote{The ranking refers to the time of the submission: 
\hao{12th June}, 2020} on the $Car$ class of the KITTI 3D detection benchmark.

Inspired by Hough voting-based 2D object detectors, Qi et al. \cite{VoteNet} proposed VoteNet to directly vote for virtual center points of objects from point clouds and to generate a group of high-quality 3D object proposals by aggregating vote features. VoteNet significantly outperforms previous approaches using only geometric information, and achieves the state-of-the-art performance on two large indoor benchmarks (i.e., ScanNet \cite{ScanNet} and SUN RGB-D \cite{SunRGBD}). However, the prediction of virtual center point is unstable for a partially occluded object. Further, Feng et al. \cite{VoteNet2} added an auxiliary branch of direction vectors to improve the prediction accuracy of virtual center points and 3D candidate boxes. In addition, a 3D object-object relationship graph between proposals is built to emphasize useful features for accurate object detection. \hao{Qi et al. \cite{ImVoteNet} proposed an ImVoteNet detector by fusing 2D object detection cues (e.g., geometric and semantic/texture cues) into a 3D voting pipeline.}
Inspired by the observation that the ground truth boxes of 3D objects provide accurate locations of intra-object parts, Shi et al. \cite{PartA2} proposed  the Part-$A^{2}$ Net, which is composed of a part-aware stage and a part-aggregation stage. The part-aware stage applies a UNet-like \cite{ronneberger2015u} network with sparse convolution and sparse deconvolution to learn point-wise features for the prediction and coarse generation of intra-object part locations. The part-aggregation stage adopts RoI-aware pooling to aggregate predicted part locations for box refinement.

\subsubsection{\bf Single Shot Methods}

These methods directly predict class probabilities and regress 3D bounding boxes of objects using a single-stage network. They do not need region proposal generation and post-processing. As a result, they can run at a high speed.
According to the type of input data, single shot methods can be divided into three categories: BEV-based, \hao{discretization-based and point-based methods.}

\textbf{BEV-based Methods.} 
These methods mainly take BEV representation as their input. Yang et al. \cite{PIXOR} discretized the point cloud of a scene with equally spaced cells and encoded the reflectance in a similar way, resulting in a regular representation. A Fully Convolution Network (FCN) network is then applied to estimate the locations and heading angles of objects. This method outperforms most single shot methods (including VeloFCN \cite{VeloFCN}, 3D-FCN \cite{3DFCN} and Vote3Deep \cite{Vote3deep}) while running at 28.6 fps. Later, Yang et al. \cite{HDNET} exploited the geometric and semantic prior information provided by High-Definition (HD) maps to improve the robustness and detection performance of \cite{PIXOR}. Specifically, they obtained the coordinates of ground points from the HD map and then used the distance relative to the ground for BEV representation 
to remedy the translation variance caused by the slope of the road. In addition, they concatenated a binary road mask with the BEV representation along the channel dimension to focus on moving objects. Since HD maps are not available everywhere, they also proposed an online map prediction module to estimate the map priors from single LiDAR point cloud. This map-aware method significantly outperforms its baseline on the TOR4D \cite{PIXOR,FastandFurious} and KITTI \cite{KITTI} datasets. However, its generalization performance to point clouds with different densities is poor. To solve this problem, Beltr{\'a}n et al. \cite{BirdNet} proposed a normalization map to consider the differences among different LiDAR sensors. The normalization map is a 2D grid with the same resolution as the BEV map, and it encodes the maximum number of points contained in each cell. It is shown that this normalization map significantly improves the generalization ability of BEV-based detectors.

\textbf{\hao{Discretization-based Methods.}} These methods convert a point cloud into a regular discrete representation, and then apply CNN to predict both categories and 3D boxes of objects.

Li et al. \cite{VeloFCN} proposed the first method to use a FCN for 3D object detection. They converted a point cloud into a 2D point map and used a 2D FCN to predict the bounding boxes and confidences of objects. Later, they \cite{3DFCN} discretized the point cloud into a 4D tensor with dimensions of length, width, height and channels, and extended the 2D FCN-based detection technologies to 3D domain for 3D object detection. Compared to \cite{VeloFCN}, 3D FCN-based method \cite{3DFCN} obtains a gain of over 20\% in accuracy, but inevitably costs more computing resources due to 3D convolutions and the sparsity of the data. To address the sparsity problem of voxels, Engelcke et al. \cite{Vote3deep} leveraged a feature-centric voting scheme to generate a set of votes for each non-empty voxel and to obtain the convolutional results by accumulating the votes. Its computational complexity is proportional to the number of occupied voxels. Li et al. \cite{3DBN} constructed a 3D backbone network by stacking multiple sparse 3D CNNs. This method is designed to save memory and accelerate computation by fully using the sparsity of voxels. This 3D backbone network extracts rich 3D features for object detection without introducing heavy computational burden.

Zhou et al. \cite{VoxelNet} presented a voxel-based end-to-end trainable framework VoxelNet. They partitioned a point cloud into equally spaced voxels and encoded the features within each voxel into a 4D tensor. A region proposal network is then connected to produce detection results. Although its performance is strong, this method is very slow due to the sparsity of voxels and 3D convolutions. Later, Yan et al. \cite{SECOND} used the sparse convolutional network \cite{sparseconv} to improve the inference efficiency of \cite{VoxelNet}. They also proposed a sine-error angle loss to solve the ambiguity between orientations of 0 and $\pi$. Sindagi et al. \cite{MVX-Net} extended VoxelNet by fusing image and point cloud features at early stages. Specifically, they projected non-empty voxels generated by \cite{VoxelNet} into the image and used a pre-trained network to extract image features for each projected voxel. These image features are then concatenated with voxel features to produce accurate 3D boxes. Compared to \cite{VoxelNet,SECOND}, this method can effectively exploit multi-modal information to reduce false positives and negatives. Lang et al. \cite{PointPillars} proposed a 3D object detector named PointPillars. This method leverages PointNet \cite{PointNet} to learn the feature of point clouds organized in vertical columns (Pillars) and encodes the learned features as a pesudo image. A 2D object detection pipeline is then applied to predict 3D bounding boxes. PointPillars outperforms most fusion approaches (including MV3D \cite{chen2017multi}, RoarNet \cite{Roarnet} and AVOD \cite{AVOD}) in terms of Average Precision (AP). Moreover, PointPillars can run at a speed of 62 fps on both the 3D and BEV KITTI \cite{KITTI} benchmarks, making it highly suitable for practical applications.

\hao{Inspired by the observation that partial spatial information of a point cloud is inevitably lost in progressively downscaled feature maps of existing single shot detectors, He et al. \cite{SA-SSD} proposed a SA-SSD detector to leverage the fine-grained structure information to improve localization accuracy. Specifically, they first converted a point cloud to a tensor and fed it into a backbone network to extract multi-stage features. In addition, an auxiliary network with point-level supervision is employed to guide the features to learn the structure of point clouds. Experimental results show that SA-SSD ranks the first\footnote{\hao{The ranking refers to the time of the submission: 
12th June, 2020}} on the $Car$ class of the KITTI BEV detection benchmark.}

\textbf{\hao{Point-based Methods.}} \hao{These methods directly take raw point clouds as their inputs. 3DSSD \cite{3DSSD} is a pioneering work in this direction. It introduces a fusion sampling strategy for Distance-FPS (D-FPS) and Feature-FPS (F-FPS) to remove time-consuming Feature Propagation (FP) layers and the refinement module in \cite{PointRCNN}. Then, a Candidate Generation (CG) layer is used to fully exploit representative points, which are further fed into an anchor-free regression head with a 3D centerness label to predict 3D object boxes. Experimental results show that 3DSSD outperforms the two-stage point-based method PointRCNN \cite{PointRCNN} while maintaining a speed of 25 fps.} 

\textbf{Other Methods.} Meyer et al. \cite{LaserNet} proposed an efficient 3D object detector called LaserNet. This method predicts a probability distribution over bounding boxes for each point and then combines these per-point distributions to generate final 3D object boxes. Further, the dense Range View (RV) representation of point cloud is used as input and a fast mean-shift algorithm is proposed to reduce the noise produced by per-point prediction. LaserNet achieves the state-of-the-art performance at the range of 0 to 50 meters, and its runtime is significantly lower than existing methods. Meyer et al. \cite{LaserNet++} then extended LaserNet \cite{LaserNet} to exploit the dense texture provided by RGB images (e.g., 50 to 70 meters). Specifically, they associated LiDAR points with image pixels by projecting 3D point clouds onto 2D images and exploited this association to fuse RGB information into 3D points. They also considered 3D semantic segmentation as an auxiliary task to learn better representations. This method achieves a significant improvement in both long-range (e.g., 50 to 70 meters) object detection and semantic segmentation while maintaining high efficiency of LaserNet. 
Inspired by the observation that points on an isolated object part can provide abundant information about position and orientation of the object, Chen et al. \cite{OHS} proposed a novel $Hotspot$ representation and the first hotspot-based anchor-free detector. Specifically, raw point clouds are first voxelized and then fed into a backbone network to produce 3D feature maps. These feature maps are used to classify hotspots and predict 3D bounding boxes simultaneously. Note that, hotspots are assigned at the last convolutional layer of the backbone network. Experimental results show that this method achieves comparable performance and is robust to sparse point clouds.
\hao{Shi et el. \cite{PointGNN} proposed a graph neural network Point-GNN to detect 3D objects from lidar point clouds. They first encoded an input point cloud as a graph of near neighbors with a fixed radius and then fed the graph into Point-GNN to predict both the categories and boxes of objects.}

\begin{table*}[t]
\centering
\caption{Comparative 3D object detection results on the KITTI test BEV detection benchmark. 3D bounding box IoU threshold is 0.7 for cars and 0.5 for pedestrians and cyclists. The modalities are LiDAR (L) and image (I). `E', `M' and `H' represent easy, moderate and hard classes of objects, respectively. For simplicity, we omit the `\%' after the value. The symbol `-' means the results are unavailable.}
\label{Tab:KITTIBEV}
\resizebox{\textwidth}{!}{
\begin{tabular}{|c|c|r|c|c|ccc|ccc|ccc|}
\hline
 \multicolumn{3}{|c|}{\multirow{2}{*}{\bf Method}} & \multirow{2}{*}{\bf Modality} & \multirow{2}{*}{\bf \tabincell{c}{Speed \\ (fps)}} & \multicolumn{3}{|c|}{\bf Cars} & \multicolumn{3}{|c|}{\bf Pedestrians} & \multicolumn{3}{|c|}{\bf Cyclists} \\\cline{6-14}
 \multicolumn{3}{|c|}{} & & & {\bf E} & {\bf M} & {\bf H} & {\bf E} & {\bf M} & {\bf H} & {\bf E} & {\bf M} & {\bf H} \\\hline
\multirow{19}{*}{\tabincell{c}{Region \\ Proposal \\-based \\ Methods}} & \multirow{6}{*}{\tabincell{c}{Multi-view \\ Methods}} & {MV3D \cite{chen2017multi}} & {L \& I} & 2.8 & 86.62 & 78.93 & 69.80 & - & - & - & - & - & - \\
  & & {AVOD \cite{AVOD}} & {L \& I} & 12.5 & 89.75 & 84.95 & 78.32 & 42.58 & 33.57 & 30.14 & 64.11 & 48.15 & 42.37 \\
  & & {ContFuse \cite{ContFuse}} & {L \& I} & 16.7 & 94.07 & 85.35 & 75.88 & - & - & - & - & - & - \\
  & & {MMF \cite{MMF}} & {L \& I} & 12.5 & 93.67 & 88.21 & 81.99 & - & - & - & - & - & - \\
  & & {SCANet \cite{Scanet}} & {L \& I} & 11.1 & 90.33 & 82.85 & 76.06 & - & - & - & - & - & - \\
  & & {RT3D \cite{RT3D}} & {L \& I} & 11.1 & 56.44 & 44.00 & 42.34 & - & - & - & - & - & - \\\cline{2-14}
  & \multirow{3}{*}{\tabincell{c}{Segmentation \\ -based \\ Methods}} & {IPOD \cite{IPOD}} & {L \& I} & 5.0 & 89.64 & 84.62 & 79.96 & 60.88 & 49.79 & 45.43 & 78.19 & 59.40 & 51.38 \\
  & & {PointRCNN \cite{PointRCNN}} & {L} & 10.0 & 92.13 & 87.39 & 82.72 & 54.77 & 46.13 & 42.84 & 82.56 & 67.24 & 60.28 \\
  & & {PointRGCN \cite{PointRGCN}} & {L} & 3.8 & 91.63 & 87.49 & 80.73 & - & - & - & - & - & - \\
  & & {PointPainting \cite{PointPainting}} & {L \& I} & 2.5 & 92.45 & 88.11 & 83.36 & 58.70 & 49.93 & 46.29 & 83.91 & 71.54 & 62.97 \\ 
  & & {STD \cite{STD}} & {L} & 12.5 & 94.74 & 89.19 & 86.42 & 60.02 & 48.72  44.55 & 81.36 & 67.23 & 59.35 \\\cline{2-14}
  & \multirow{6}{*}{\tabincell{c}{Frustum \\ -based \\ Methods}} & {F-PointNets \cite{F-PointNet}} & {L \& I} & 5.9 & 91.17 & 84.67 & 74.77 & 57.13 & 49.57 & 45.48 & 77.26 & 61.37 & 53.78 \\
  & & {SIFRNet \cite{SIFRNet}} & {L \& I} & - & - & - & - & - & - & - & - & - & -\\
  & & {PointFusion \cite{PointFusion}} & {L \& I} & - & - & - & - & - & - & - & - & - & - \\
  & & {RoarNet \cite{Roarnet}} & {L \& I} & 10.0 & 88.20 & 79.41 & 70.02 & - & - & - & - & - & - \\
  & & {F-ConvNet \cite{F-ConvNet}} & {L \& I} & 2.1 & 91.51 & 85.84 & 76.11 & 57.04 & 48.96 & 44.33 & 84.16 & 68.88 & 60.05 \\
  & & {\tabincell{c}{Patch Refinement \cite{Patch}}} & {L} & 6.7 & 92.72 & 88.39 & 83.19 & - & - & - & - & - & -\\\cline{2-14}
  & \multirow{4}{*}{\tabincell{c}{Other \\ Methods}} & {3D IoU loss \cite{3D-IoU}} & {L} & 12.5 & 91.36 & 86.22 & 81.20 & - & - & - & - & - & - \\
  & & {\tabincell{c}{Fast Point R-CNN \cite{Fast-PointRCNN}}} & {L} & 16.7 & 90.76 & 85.61 & 79.99 & - & - & - & - & - & - \\
  & & {PV-RCNN \cite{PVRCNN}} & {L} & 12.5 & 94.98 & 90.65 & 86.14 & - & - & - & 82.49 & 68.89 & 62.41 \\
  & & {VoteNet \cite{VoteNet}} & {L} & - & - & - & - & - & - & - & - & - & - \\
  & & {Feng et al. \cite{VoteNet2}} & {L} & - & - & - & - & - & - & - & - & - & - \\
  & & {ImVoteNet \cite{ImVoteNet}} & {L \& I} & - & - & - & - & - & - & - & - & - & - \\
  & & {Part-A\^{}2 \cite{PartA2}} & {L} & 12.5 & 91.70 & 87.79 & 84.61 & - & - & - & 81.91 & 68.12 & 61.92 \\\hline
  \multirow{13}{*}{\tabincell{c}{Single \\ Shot \\ Methods}} & \multirow{3}{*}{\tabincell{c}{BEV-based \\ Methods}} & {PIXOR \cite{PIXOR}} & {L} & 28.6 & 83.97 & 80.01 & 74.31 & - & - & - & - & - & - \\
  & & {HDNET \cite{HDNET}} & {L} & 20.0 & 89.14 & 86.57 & 78.32 & - & - & - & - & - & - \\
  & & {BirdNet \cite{BirdNet}} & {L} & 9.1 & 76.88 & 51.51 & 50.27 & 20.73 & 15.80 & 14.59 & 36.01 & 23.78 & 21.09 \\\cline{2-14}
  & \multirow{8}{*}{\tabincell{c}{Discretization \\ -based \\ Methods}} &  {VeloFCN \cite{VeloFCN}} & {L} & 1.0 & 0.02 & 0.14 & 0.21 & - & - & - & - & - & - \\
  & & {3D FCN \cite{3DFCN}} & {L} & {\textless0.2} & 70.62 & 61.67 & 55.61 & - & - & - & - & - & - \\
  & & {Vote3Deep \cite{Vote3deep}} & {L} & - & - & - & - & - & - & - & - & - & - \\
  & & {3DBN \cite{3DBN}} & {L} & 7.7 & 89.66 & 83.94 & 76.50 & - & - & - & - & - & -\\
  & & {VoxelNet \cite{VoxelNet}} & {L} & 2.0 & 89.35 & 79.26 & 77.39 & 46.13 & 40.74 & 38.11 & 66.70 & 54.76 & 50.55 \\
  & & {SECOND \cite{SECOND}} & {L} & 26.3 & 89.39 & 83.77 & 78.59 & 55.99 & 45.02 & 40.93 & 76.50 & 56.05 & 49.45 \\
  & & {MVX-Net \cite{MVX-Net}} & {L \& I} & 16.7 & 92.13 & 86.05 & 78.68 & - & - & - & - & - & - \\
  & & {PointPillars \cite{PointPillars}} & {L} & 62.0 & 90.07 & 86.56 & 82.81 & 57.60 & 48.64 & 45.78 & 79.90 & 62.73 & 55.58 \\
  & & {SA-SSD \cite{SA-SSD}} & {L} & 25.0 & 95.03 & 91.03 &	85.96 & - & - & - & - & - & - \\\cline{2-14}
  & \tabincell{c}{Point-based \\ Methods} & {3DSSD \cite{3DSSD}} & {L} & 25.0 & 92.66 & 89.02 & 85.86 & 60.54 & 49.94 & 45.73 & 85.04 & 67.62 & 61.14 \\\hline
  & \multirow{2}{*}{\tabincell{c}{Other \\ Methods}} & {LaserNet \cite{LaserNet}} & {L} & 83.3 & 79.19 & 74.52 & 68.45 & - & - & - & - & - & - \\
  & & {LaserNet++ \cite{LaserNet++}} & {L \& I} & 26.3 & - & - & - & - & - & - & - & - & - \\
  & & {OHS-Dense \cite{OHS}} & {L} & 33.3 & 93.73 & 88.11 & 84.98 & 50.87 & 44.59 & 42.14 & 82.13 & 66.86 & 60.86 \\
  & & {OHS-Direct \cite{OHS}} & {L} & 33.3 & 93.59 & 87.95 & 83.21 & 55.90 & 49.48 & 45.79 & 79.66 & 67.20 & 61.04 \\
  & & {Point-GNN \cite{PointGNN}} & {L} & 1.7 & 93.11 & 89.17 & 83.90 & 55.36 & 47.07 & 44.61 & 81.17 & 67.28 & 59.67\\\hline
\end{tabular}}
\end{table*}

\subsection{3D Object Tracking}
Given the locations of an object in the first frame, the task of object tracking is to estimate its state in subsequent frames \cite{hucorrelation,liu2019robust}. Since 3D object tracking can use the rich geometric information in point clouds, it is expected to overcome several drawbacks faced by image-based tracking, including occlusion, illumination and scale variation. 

Inspired by the success of Siamese network \cite{SiamFC} for imaged-based object tracking, Giancola et al. \cite{3DSiamese} proposed a 3D Siamese network with shape completion regularization. 
Specifically, they first generated candidates using a Kalman filter, and encoded model and candidates into a compact representation using shape regularization. The cosine similarity is then used to search the location of the tracked object in the next frame. This method can be used as an alternative for object tracking, and significantly outperforms most 2D object tracking methods, including $\mathrm{STAPLE_{CA}}$ \cite{StapleCA} and SiamFC \cite{SiamFC}. To efficiently search the target object, Zarzar et al. \cite{3DSiameseV2} leveraged a 2D Siamese network to generate a large number of coarse object candidates on BEV representation. They then refined the candidates by exploiting the cosine similarity in 3D Siamese network. This method significantly outperforms \cite{3DSiamese} in terms of both precision (i.e., by 18\%) and success rate (i.e., by 12\%). Simon et al. \cite{Complexer-YOLO-tracking} proposed a 3D object detection and tracking architecture for semantic point clouds. They first generated voxelized semantic point clouds by fusing 2D visual semantic information, and then utilized the temporal information to improve accuracy and robustness of multi-target tracking. In addition, they introduced a powerful and simplified evaluation metric (i.e., Scale-Rotation-Translation score (SRFs)) to speed up training and inference. Complexer-YOLO achieves promising tracking performance and can still run in real-time. \hao{Further, Qi et al. \cite{P2B} proposed a Point-to-Box (P2B) network. They fed template and search areas into the backbone to obtain their seeds. The search area seeds are augmented with target-specific features and then the potential target centers are regressed by Hough voting. Experimental results show that P2B outperforms \cite{3DSiamese} by over 10\% while running at 40 fps.}

\begin{figure}
\centering
\includegraphics[width=\columnwidth,keepaspectratio]{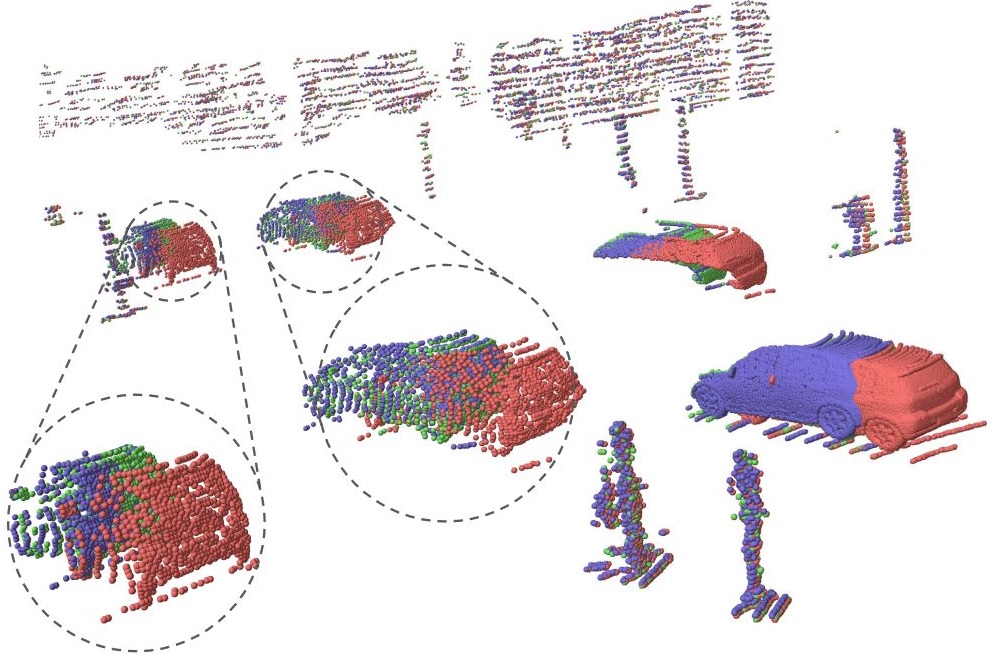}
\caption{A 3D scene flow between two KITTI point clouds, originally shown in \cite{liu2019flownet3d}. Point clouds $\mathcal{X}$, $\mathcal{Y}$ and the translated point cloud of $\mathcal{X}$ are highlighted in red, green, and blue, respectively.}
\label{fig:scene-flow}
\end{figure}

\subsection{3D Scene Flow Estimation}
\hao{Given two point clouds $\mathcal{X}$ and $\mathcal{Y}$, 3D scene flow $D=\{d_i\}^N$ describes the movement of each point $x_i$ in $\mathcal{X}$ to its corresponding position $x_i'$ in $\mathcal{Y}$, such that $x_i'=x_i+d_i$. Figure \ref{fig:scene-flow} shows a 3D scene flow between two KITTI point clouds.} Analogous to optical flow estimation in 2D vision, several methods have started to learn useful information (e.g. 3D scene flow, spatial-temporary information) from a sequence of point clouds.

Liu et al. \cite{liu2019flownet3d} proposed FlowNet3D to directly learn scene flows from a pair of consecutive point clouds. FlowNet3D learns both point-level features and motion features through a flow embedding layer. However, there are two problems with FlowNet3D. First, some predicted motion vectors differ significantly from the ground truth in their directions. Second, it is difficult to apply FlowNet to non-static scenes, especially for the scenes which are dominated by deformable objects. To solve this problem, Wang et al. \cite{FlowNet3D++} introduced a cosine distance loss to minimize the angle between the predictions and the ground truth. In addition, they also proposed a point-to-plane distance loss to improve the accuracy for both rigid and dynamic scenes. Experimental results show that these two loss terms improve the accuracy of FlowNet3D from 57.85\% to 63.43\%, and speed up and stabilize the training process. Gu et al. \cite{gu2019hplflownet} proposed a Hierarchical Permutohedral Lattice FlowNet (HPLFlowNet) to directly estimate scene flow from large-scale point clouds. Several bilateral convolution layers are proposed to restore structural information from raw point clouds, while reducing the computational cost.

To effectively process sequential point clouds, Fan and Yang \cite{fan2019pointrnn} proposed PointRNN, PointGRU and PointLSTM networks and a sequence-to-sequence model to track moving points. PointRNN, PointGRU, and PointLSTM are able to capture the spatial-temporary information and model dynamic point clouds. Similarly, Liu et al. \cite{MeteorNet} proposed MeteorNet to directly learn a representation from dynamic point clouds. This method learns to aggregate information from spatiotemporal neighboring points. Direct grouping and chained-flow grouping are further introduced to determine the temporal neighbors. However, the performance of the aforementioned methods is limited by the scale of datasets. Mittal et al. \cite{Self-supervised} proposed two self-supervised losses to train their network on large unlabeled datasets. Their main idea is that a robust scene flow estimation method should be effective in both forward and backward predictions. Due to the unavailability of scene flow annotation, the nearest neighbor of the predicted transformed point is considered as pesudo ground truth. However, the true ground truth may not be the same as the nearest point. To avoid this problem, they computed the scene flow in the reverse direction and proposed a cycle consistency loss to translate the point to the original position. Experimental results show that this self-supervised method exceeds the state-of-the-art performance of supervised learning-based methods. 

\subsection{Summary}
The KITTI \cite{KITTI} benchmark is one of the most influential datasets in autonomous driving and has been commonly used in both academia and industry. Tables \ref{Tab:KITTI3D} and \ref{Tab:KITTIBEV} present the results achieved by different detectors on the KITTI test 3D benchmarks. The following observations can be made:

\begin{itemize}
\item[$\bullet$] Region proposal-based methods are the most frequently investigated methods among these two categories, and outperform single shot methods by a large margin on both KITTI test 3D and BEV benchmarks.
\item[$\bullet$] There are two limitations for existing 3D object detectors. First, the long-range detection capability of existing methods is relatively poor. Second, how to fully exploit the texture information in images is still an open problem.
\item[$\bullet$] Multi-task learning is a future direction in 3D object detection. E.g., MMF \cite{MMF} learns a cross-modality representation to achieve state-of-the-art detection performance by incorporating multiple tasks.
\item[$\bullet$] 3D object tracking and scene flow estimation are emerging research topics, and have gradually attracted increasing attention since 2019. 
\end{itemize}

\section{3D Point Cloud Segmentation} \label{sec:scene_segmentation}

3D point cloud segmentation requires the understanding of both the global geometric structure and the fine-grained details of each point. According to the segmentation granularity, 3D point cloud segmentation methods can be classified into three categories: \textit{semantic segmentation} (scene level), \textit{instance segmentation} (object level) and \textit{part segmentation} (part level).

\begin{figure*}
\centering
\includegraphics[width=1\textwidth]{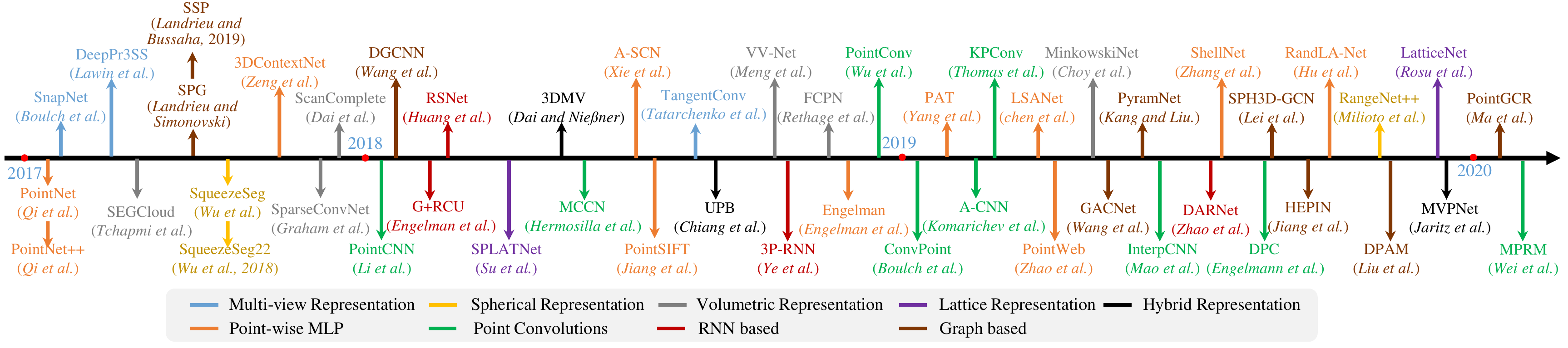}
\caption{Chronological overview of the most relevant deep learning-based 3D semantic segmentation methods. }
\label{fig:milestone_sem}
\end{figure*}

\subsection{3D Semantic Segmentation}
Given a point cloud, the goal of semantic segmentation is to separate it into several subsets according to the semantic meanings of points. Similar to the taxonomy for 3D shape classification (Section \ref{sec:shape_classification}), there are \qy{four paradigms for semantic segmentation: projection-based, discretization-based,  point-based, and hybrid methods.}

\begin{figure}[bht]
\centering
\includegraphics[width=0.45\textwidth]{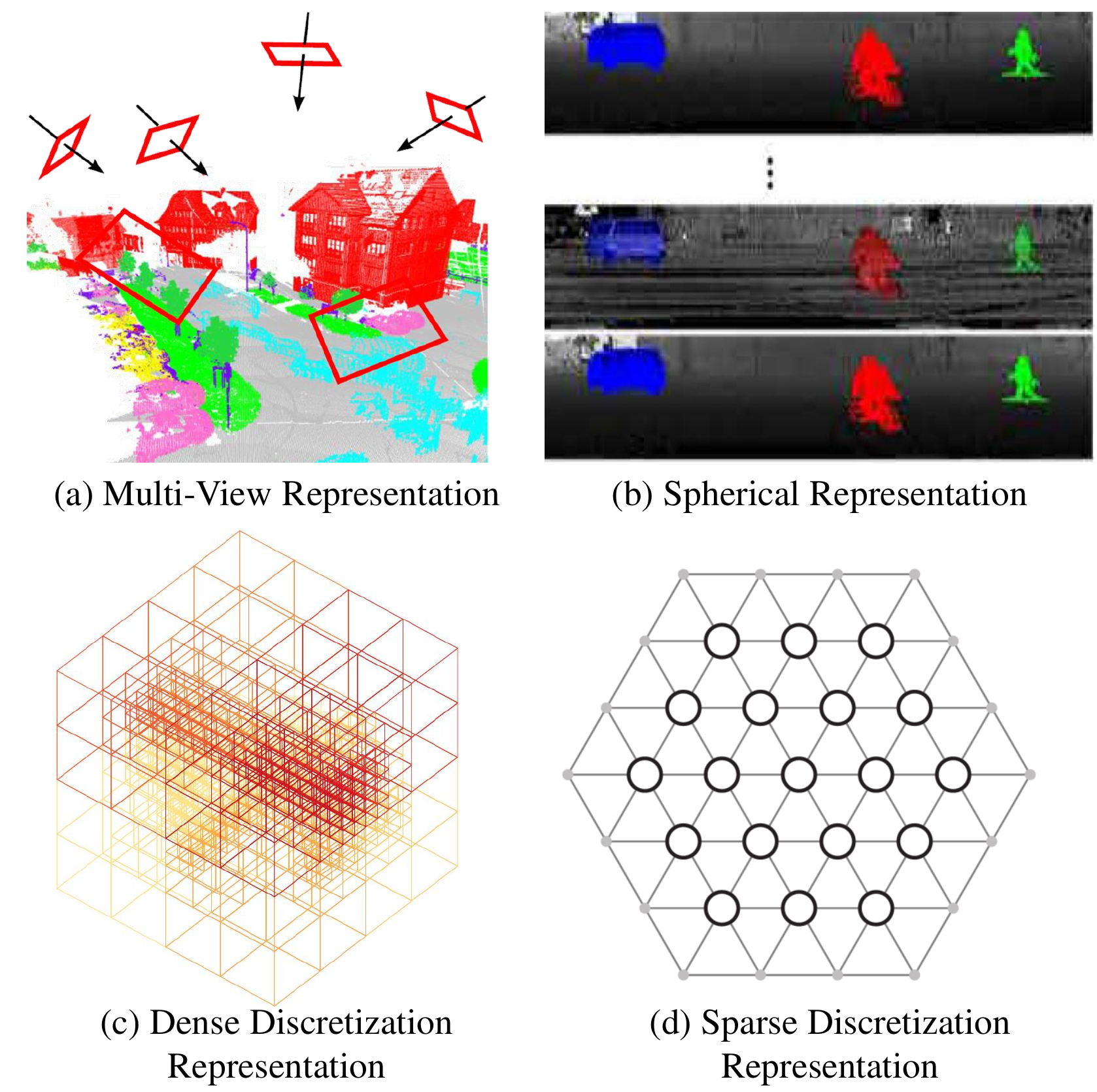}
\caption{An illustration of the intermediate representation. (a) and (b) are originally shown in \cite{snapnet} and \cite{wu2018squeezeseg}, respectively. \label{fig:intermediate_representation}}
\end{figure}

\qy{The first step of both the projection and discretization-based methods is to transform a point cloud to an intermediate regular representation, such as multi-view  \cite{projective, snapnet}, spherical  \cite{wu2018squeezeseg, wu2019squeezesegv2,milioto2019rangenet++}, volumetric  \cite{VVNET, rethage2018fully, sparseconv}, permutohedral lattice  \cite{su2018splatnet, rosu2019latticenet}, and hybrid representations \cite{3DMV, jaritz2019multi}, as shown in Fig. \ref{fig:intermediate_representation}. The intermediate segmentation results are then projected back to the raw point cloud. In contrast, point-based methods directly work on irregular point clouds. Several representative methods are shown in Fig. \ref{fig:milestone_sem}.}

\subsubsection{\bf{\qy{Projection-based Methods}}}
\qy{These methods usually project a 3D point cloud into 2D images, including multi-view and spherical images. }

\textbf{Multi-view Representation.} 
Lawin et al. \cite{projective} first projected a 3D point cloud onto 2D planes from multiple virtual camera views. Then, a multi-stream FCN is used to predict pixel-wise scores on  synthetic images. The final semantic label of each point is obtained by fusing the re-projected scores over different views. Similarly, Boulch et al. \cite{snapnet} first generated several RGB and depth snapshots of a point cloud using multiple camera positions. They then performed pixel-wise labeling on these snapshots using 2D segmentation networks. The scores predicted from RGB and depth images are further fused using residual correction \cite{residual_correction}. Based on the assumption that point clouds are sampled from locally Euclidean surfaces, Tatarchenko et al. \cite{tatarchenko2018tangent} introduced tangent convolutions for dense point cloud segmentation. This method first projects the local surface geometry around each point to a virtual tangent plane. Tangent convolutions are then directly operated on the surface geometry. This method shows great scalability and is able to process large-scale point clouds with millions of points. Overall, the performance of multi-view segmentation methods is sensitive to viewpoint selection and occlusions. Besides, these methods have not fully exploited the underlying geometric and structural information, as the projection step inevitably introduces information loss. 

\textbf{Spherical Representation.} 
To achieve fast and accurate segmentation of 3D point clouds,  Wu et al. \cite{wu2018squeezeseg} proposed an end-to-end network based on SqueezeNet \cite{iandola2016squeezenet} and Conditional Random Field (CRF). To further improve segmentation accuracy, SqueezeSegV2 \cite{wu2019squeezesegv2} is introduced to address domain shift by utilizing an unsupervised domain adaptation pipeline. Milioto et al. \cite{milioto2019rangenet++} proposed RangeNet++ for real-time semantic segmentation of LiDAR point clouds. The semantic labels of 2D range images are first transferred to 3D point clouds, an efficient GPU-enabled KNN-based post-processing step is further used to alleviate the problem of discretization errors and blurry inference outputs.
Compared to single view projection, spherical projection retains more information and is suitable for the labeling of LiDAR point clouds. However, this intermediate representation inevitably brings several problems such as discretization errors and occlusions.

\subsubsection{\bf{\qy{Discretization-based Methods}} }
\qy{These methods usually convert a point cloud into a dense/sparse discrete representation, such as  volumetric and sparse permutohedral lattices.}

\textbf{Dense Discretization Representation.}
Early methods usually voxelized the point clouds as dense grids and then leverage the standard 3D convolutions. Huang et al. \cite{huang2016point} first divided a point cloud into a set of occupancy voxels, then fed these intermediate data to a fully-3D CNN for voxel-wise segmentation. Finally, all points within a voxel are assigned the same semantic label as the voxel. The performance of this method is severely limited by the granularity of the voxels and the boundary artifacts caused by the point cloud partition. Further, Tchapmi et al. \cite{segcloud} proposed SEGCloud to achieve fine-grained and global consistent semantic segmentation. This method introduces a deterministic trilinear interpolation to map the coarse voxel predictions generated by 3D-FCNN \cite{FCNN} back to the point cloud, and then uses Fully Connected CRF (FC-CRF) to enforce spatial consistency of these inferred per-point labels. Meng et al. \cite{VVNET} introduced a kernel-based interpolated variational autoencoder architecture to encode the local geometrical structures within each voxel. Instead of a binary occupancy representation, RBFs are employed for each voxel to obtain a continuous representation and capture the distribution of points in each voxel. VAE is further used to map the point distribution within each voxel to a compact latent space. Then, both symmetry groups and an equivalence CNN are used to achieve robust feature learning. 

\qy{Thanks to the good scalability of 3D CNN, volumetric-based networks are free to be trained and tested on point clouds with different spatial sizes.}
In Fully-Convolutional Point Network (FCPN) \cite{rethage2018fully}, different levels of geometric relations are first hierarchically abstracted from point clouds, 3D convolutions and weighted average pooling are then used to extract features and incorporate long-range dependencies. This method can process large-scale point clouds and has good scalability during inference. Dai et al. \cite{scancomplete} proposed ScanComplete to achieve 3D scan completion and per-voxel semantic labeling. This method leverages the scalability of fully-convolutional neural networks and can adapt to different input data sizes during training and test. A coarse-to-fine strategy is used to hierarchically improve the resolution of the predicted results.

Overall, the volumetric representation naturally preserves the neighborhood structure of 3D point clouds. Its regular data format also allows direct application of standard 3D convolutions. These factors lead to a steady performance improvement in this area. However, the voxelization step inherently introduces discretization artifacts and information loss. Usually, a high resolution leads to high memory and computational costs, while a low resolution introduces loss of details. It is non-trivial to select an appropriate grid resolution in practice. 

\textbf{Sparse Discretization Representation.}
Volumetric representation is naturally sparse, as the number of non-zero values only accounts for a small percentage. Therefore, it is inefficient to apply dense convolution neural networks on the spatially-sparse data. To this end, Graham et al. \cite{sparseconv} proposed submanifold sparse convolutional networks based on the indexing structure. This method significantly reduces memory and computational costs by restricting the output of convolution to be only related to occupied voxels. Meanwhile, its sparse convolution can also control the sparsity of the extracted features. This submanifold sparse convolution is suitable for efficient processing of high-dimensional and spatially-sparse data. Further, Choy et al. \cite{Minkowski} proposed a 4D spatio-temporal convolutional neural network called MinkowskiNet for 3D video perception. A generalized sparse convolution is proposed to effectively process high-dimensional data. A trilateral-stationary conditional random field is further applied to enforce consistency.

On the other hand, Su et al. \cite{su2018splatnet} proposed the Sparse Lattice Networks (SPLATNet) based on Bilateral Convolution Layers  (BCLs). This method first interpolates a raw point cloud to a permutohedral sparse lattice, BCL is then applied to convolve on occupied parts of the sparsely populated lattice. The filtered output is then interpolated back to the raw point cloud. In addition, this method allows flexible joint processing of multi-view images and point clouds. Further, Rosu et al. \cite{rosu2019latticenet} proposed LatticeNet to achieve efficient processing of large point clouds. A data-dependent interpolation module called DeformsSlice is also introduced to back project the lattice feature to point clouds.



\subsubsection{\bf{Hybrid Methods}}
To further leverage all available information, several methods have been proposed to learn multi-modal features from 3D scans. Dai and Nie{\ss}ner \cite{3DMV} \why{presented} a joint 3D-multi-view network to combine RGB features and geometric features. A 3D CNN stream and several 2D streams are used to extract features, and a differentiable back-projection layer is proposed to jointly fuse the learned 2D embeddings and 3D geometric features. Further, Chiang et al. \cite{jointpoint-based} proposed a unified point-based framework to learn 2D textural appearance, 3D structures and global context features from point clouds. This method directly applies point-based networks to extracts local geometric features and global context from sparsely sampled point sets without any voxelization. Jaritz et al. \cite{jaritz2019multi} proposed Multi-view PointNet (MVPNet) to aggregate appearance features from 2D multi-view images and spatial geometric features in the canonical point cloud space.

\subsubsection{\bf{Point-based Methods}}
Point-based networks directly work on irregular point clouds. However, point clouds are orderless and unstructured, making it infeasible to directly apply standard CNNs. To this end, the pioneering work PointNet \cite{PointNet} is proposed to learn per-point features using shared MLPs and global features using symmetrical pooling functions. Based on PointNet, a series of point-based networks have been proposed recently. Overall, these methods can be roughly divided into pointwise MLP methods, point convolution methods, RNN-based methods, and graph-based methods.

\textbf{Pointwise MLP Methods.} 
These methods usually use shared MLP as the basic unit in their network for its high efficiency. However, point-wise features extracted by  shared MLP cannot capture the local geometry in point clouds and the mutual interactions between points \cite{PointNet}. To capture wider context for each point and learn richer local structures, several dedicated networks have been introduced, including methods based on neighboring feature pooling, attention-based aggregation, and local-global feature concatenation.

\begin{figure}[t]
\centering
\includegraphics[width=0.45\textwidth]{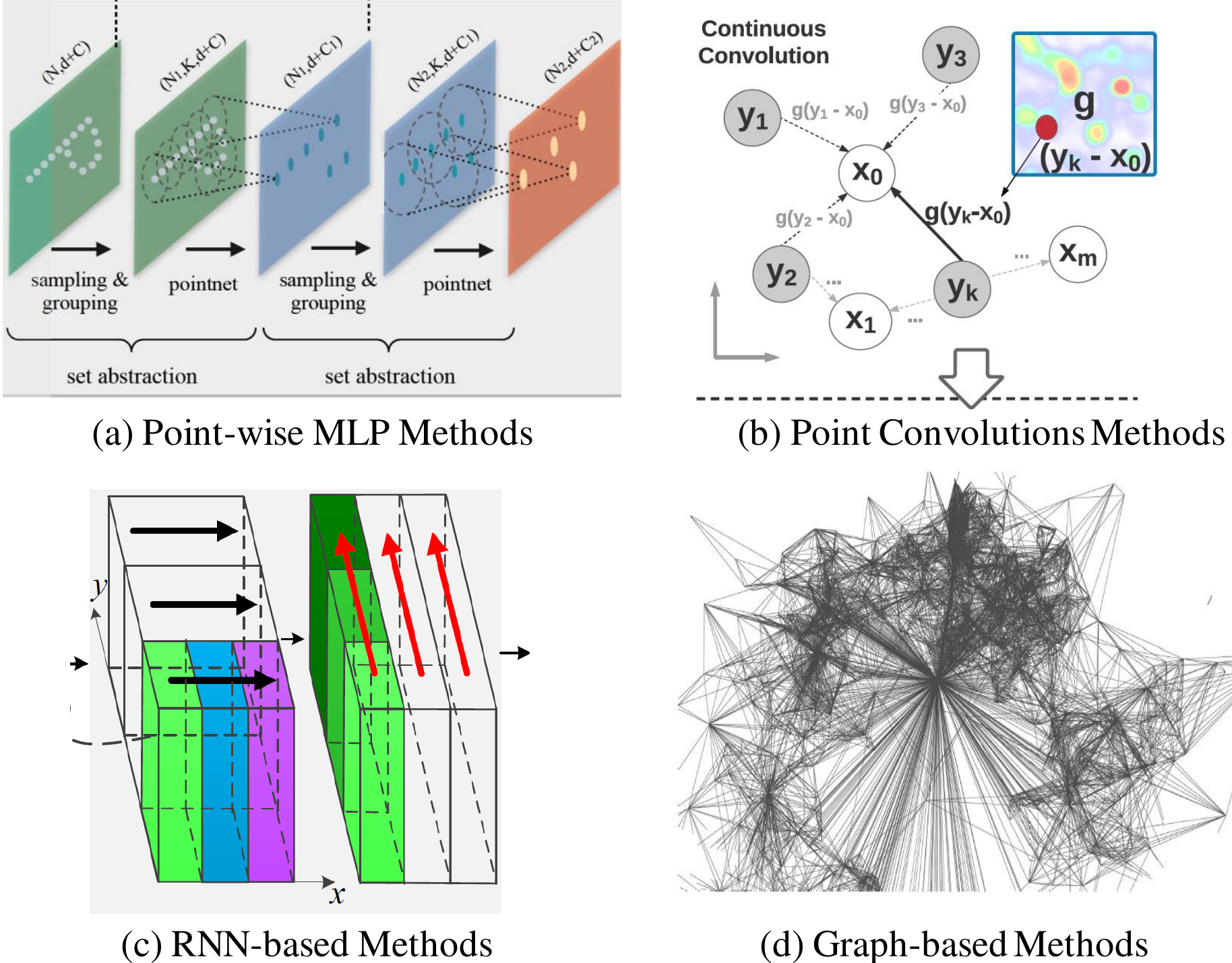}
\caption{An illustration of point-based methods. (a)-(d) are originally shown in \cite{PointNet++, PCCN, 3P-RNN, SPGraph}, respectively.}
\label{fig:Pointnet++}
\end{figure}

\textit{Neighboring feature pooling:} 
To capture local geometric patterns, these methods learn a feature for each point by aggregating the information from local neighboring points. In particular, PointNet++ \cite{PointNet++} groups points hierarchically and progressively learns from larger local regions, as illustrated in Fig. \ref{fig:Pointnet++}(a). Multi-scale grouping and multi-resolution grouping are also proposed to overcome the problems caused by non-uniformity and varying density of point clouds. Later, Jiang et al. \cite{PointSIFT} proposed a PointSIFT module to achieve orientation encoding and scale awareness. This module stacks and encodes the information from eight spatial orientations through a three-stage ordered convolution. Multi-scale features are concatenated to achieve adaptivity to different scales. Different from the grouping techniques used in PointNet++ (i.e., ball query), \why{Engelmann} et al. \cite{Know} utilized $K$-means clustering and KNN to separately define two neighborhoods in the world space and feature space. Based on the assumption that points from the same class are expected to be closer in feature space, a pairwise distance loss and a centroid loss are introduced to further regularize feature learning. To model the mutual interactions between different points, Zhao et al. \cite{PointWeb2019} proposed PointWeb to explore the relations between all pairs of points in a local region by densely constructing a locally fully-linked web. 
An Adaptive Feature Adjustment (AFA) module is proposed to achieve information interchange and feature refinement. This aggregation operation helps the network to learn a discriminative feature representation. Zhang et al. \cite{zhang2019shellnet} proposed a permutation invariant convolution called Shellconv based on the statistics from concentric spherical shells. This method first queries a set of multi-scale concentric spheres, the max-pooling operation is then used within different shells to summarize the statistics, MLPs and 1D convolution are used to obtain the final convolution output. Hu et al. \cite{randla} proposed an efficient and lightweight network called RandLA-Net for large-scale point cloud segmentation. This network utilizes random point sampling to achieve remarkably high efficiency in terms of memory and computation. A local feature aggregation module is further proposed to capture and preserve geometric features. 

\textit{Attention-based aggregation:} To further improve segmentation accuracy, an attention mechanism \cite{all_you_need} is introduced to point cloud segmentation. Yang et al. \cite{Yang2019} 
proposed a group shuffle attention to model the relations between points, and presented a permutation-invariant, task-agnostic and differentiable Gumbel Subset Sampling (GSS) to replace the widely used FPS approach. This module is less sensitive to outliers and can select a representative subset of points. To better capture the spatial distribution of a point cloud, Chen et al. \cite{LSANet} proposed a Local Spatial Aware (LSA) layer to learn spatial awareness weights based on the spatial layouts and the local structures of point clouds. 
Similar to CRF, Zhao et al. \cite{zhao2019pooling} proposed an Attention-based Score Refinement (ASR) module to post-process the segmentation results produced by the network. The initial segmentation result is refined by pooling the scores of neighboring points with learned attention weights. This module can be easily integrated into existing deep networks to improve segmentation performance. 

\textit{Local-global concatenation:} Zhao et al. \cite{Zhao2019} proposed a permutation-invariant PS${^2}$-Net to incorporate local structures and global context from point clouds. Edgeconv \cite{Wang2019} and NetVLAD \cite{netvlad} are repeatedly stacked to capture the local information and scene-level global features.

\begin{table*}[]
\centering
\caption{Comparative semantic segmentation results on the S3DIS (including both Area5 and 6-fold cross validation) \cite{S3DIS}, Semantic3D (including both \textit{semantic-8} and \textit{reduced-8} subsets) \cite{Semantic3d} , ScanNet \cite{ScanNet}, and SemanticKITTI \cite{semantickitti} datasets. Overall Accuracy (OA), Mean Intersection over Union (mIoU) are the main evaluation metric. For simplicity, we omit the `\%' after the value. The symbol `-' means the results are unavailable.}
\label{tab:segmentation_results}
\resizebox{\textwidth}{!}{%
\begin{tabular}{|c|c|r|c|c|c|c|c|c|c|c|c|c|c|}
\hline
\multicolumn{3}{|c|}{\multirow{2}{*}{\textbf{Method}}} & \multicolumn{4}{c|}{\textbf{S3DIS}} & \multicolumn{4}{c|}{\textbf{Semantic3D}} & \multicolumn{2}{c|}{\textbf{ScanNet(v2)}} & \multirow{2}{*}{\textbf{\begin{tabular}[c]{@{}c@{}}Sem.\\ KITTI\\ (mIoU)\end{tabular}}} \\ \cline{4-13}
\multicolumn{3}{|c|}{} & \textbf{\begin{tabular}[c]{@{}c@{}}Area5\\ (OA)\end{tabular}} & \textbf{\begin{tabular}[c]{@{}c@{}}Area5\\ (mIoU)\end{tabular}} & \textbf{\begin{tabular}[c]{@{}c@{}}6-fold\\ (mIoU)\end{tabular}} & \textbf{\begin{tabular}[c]{@{}c@{}}6-fold\\ (mIoU)\end{tabular}} & \textbf{\begin{tabular}[c]{@{}c@{}}sem.\\ (OA)\end{tabular}} & \textbf{\begin{tabular}[c]{@{}c@{}}sem.\\ (mIoU)\end{tabular}} & \textbf{\begin{tabular}[c]{@{}c@{}}red.\\ (OA)\end{tabular}} & \textbf{\begin{tabular}[c]{@{}c@{}}red.\\ (mIoU)\end{tabular}} & \textbf{OA} & \textbf{mIoU} &  \\ \hline
\multirow{6}{*}{\begin{tabular}[c]{@{}c@{}}Projection\\-based \\ Methods\end{tabular}} 
& \multirow{3}{*}{Multi-view}
& DeePr3SS \cite{projective} & - & - & - & - & - & - & 88.9 & 58.5 & - & - & - \\ \cline{3-14} 
 &  & SnapNet \cite{snapnet} & - & - & - & - & 91.0 & 67.4 & 88.6 & 59.1 & - & - & - \\ \cline{3-14}
  &  & TangentConv \cite{tatarchenko2018tangent} & 82.5 & 52.8 & - & - & - & - & - & - & 80.1 & 40.9 & 40.9 \\ \cline{2-14} 
 & \multirow{3}{*}{Spherical} & SqueezeSeg \cite{wu2018squeezeseg} & - & - & - & - & - & - & - & - & - & - & 29.5 \\ \cline{3-14} 
 &  & SqueezeSegV2 \cite{wu2019squeezesegv2} & - & - & - & - & - & - & - & - & - & - & 39.7 \\ \cline{3-14} 
 &  & RangeNet++ \cite{milioto2019rangenet++} & - & - & - & - & - & - & - & - & - & - & 52.2 \\  \hline 
\multirow{6}{*}{\begin{tabular}[c]{@{}c@{}}Discretization\\-based \\ Methods\end{tabular}} 
 & \multirow{4}{*}{Volumetric} & SEGCloud \cite{segcloud} & - & 48.9 & - & - & - & - & 88.1 & 61.3 & - & - & - \\ \cline{3-14} 
 &  & SparseConvNet \cite{sparseconv} & - & - & - & - & - & - & - & - & - & 72.5 & - \\ \cline{3-14} 
 &  & MinkowskiNet \cite{Minkowski} & - & - & - & - & - & - & - & - & - & 73.6 & - \\ \cline{3-14} 
 &  & VV-Net \cite{VVNET} & - & - & 87.8 & 78.2 & - & - & - & - & - & - & - \\ \cline{2-14} 
 & \multirow{2}{*}{{\begin{tabular}[c]{@{}c@{}}Permutohedral\\lattice \end{tabular}}} 
 & SPLATNet \cite{su2018splatnet} & - & - & - & - & - & - & - & - & - & 39.3 & 18.4 \\ \cline{3-14} 
 &  & LatticeNet \cite{rosu2019latticenet} & - & - & - & - & - & - & - & - & - & 64.0 & 52.2 \\ \hline 
 \multirow{3}{*}{\begin{tabular}[c]{@{}c@{}} Hybrid \\ Methods\end{tabular}} 
 & \multirow{3}{*}{Hybrid} & 3DMV \cite{3DMV} & - & - & - & - & - & - & - & - & - & 48.4 & - \\ \cline{3-14} 
 &  & UPB \cite{jointpoint-based} & - & - & - & - & - & - & - & - & - & 63.4 & - \\ \cline{3-14} 
 &  & MVPNet \cite{jaritz2019multi} & - & - & - & - & - &  & - & - & - & 64.1 & - \\ \hline
\multirow{30}{*}{\begin{tabular}[c]{@{}c@{}}Point\\-based\\ Methods\end{tabular}} & \multirow{11}{*}{\begin{tabular}[c]{@{}c@{}}Point-wise\\ MLP\end{tabular}} & PointNet \cite{PointNet} & - & 41.1 & 78.6 & 47.6 & - & - & - & - & - & - & 14.6 \\ \cline{3-14} 
 &  & PointNet++ \cite{PointNet++} & - & - & 81.0 & 54.5 & 85.7 & 63.1 & - & - & 84.5 & 33.9 & 20.1 \\ \cline{3-14} 
 &  & PointSIFT \cite{PointSIFT} & - & - & 88.7 & 70.2 & - & - & - & - & 86.2 & 41.5 & - \\ \cline{3-14} 
 &  & Engelmann \cite{Knowneigh} & 84.2 & 52.2 & 84.0 & 58.3 & - & - & - & - & - & - & - \\ \cline{3-14} 
 &  & 3DContextNet \cite{Zeng2017} & - & - & 84.9 & 55.6 & - & - & - & - & - & - & - \\ \cline{3-14} 
 &  & A-SCN \cite{Xie2018a} & - & - & 81.6 & 52.7 & - & - & - & - & - & - & - \\ \cline{3-14} 
 &  & PointWeb \cite{PointWeb2019} & 87.0 & 60.3 & 87.3 & 66.7 & - & - & - & - & 85.9 & - & - \\ \cline{3-14} 
 &  & PAT \cite{Yang2019} &  & 60.1 &  & 64.3 & - & - & - & - & - & - & - \\ \cline{3-14} 
 &  & LSANet \cite{LSANet} & - & - & 86.8 & 62.2 & - & - & - & - & 85.1 & - & - \\ \cline{3-14} 
 &  & ShellNet \cite{zhang2019shellnet} & - & - & 87.1 & 66.8 & - & - & 93.2 & 69.3 & 85.2 & - & - \\ \cline{3-14} 
 &  & RandLA-Net \cite{randla} & - & - & 88.0 & 70.0 & 94.6 & 74.8 & 94.8 & 77.4 & - & - & 55.9 \\ \cline{2-14} 
 & \multirow{7}{*}{\begin{tabular}[c]{@{}c@{}}Point\\convolution \end{tabular}} & PointCNN \cite{li2018pointcnn} & 85.9 & 57.3 & 88.1 & 65.4 & - & - & - & - & 85.1 & 45.8 & - \\ \cline{3-14} 
 &  & PCCN \cite{PCCN} & - & 58.3 & - & - & - & - & - & - & - & - & - \\ \cline{3-14} 
 &  & A-CNN \cite{Komarichev2019} & - & - & 87.3 & - & - & - & - & - & 85.4 & - & - \\ \cline{3-14} 
 &  & ConvPoint \cite{Boulch2019ConvPoint} & - & - & 88.8 & 68.2 & 93.4 & 76.5 & - & - & - & - & - \\ \cline{3-14} 
 &  & KPConv \cite{kpconv} & - & 67.1 & - & 70.6 & - & - & 92.9 & 74.6 & - & 68.4 & - \\ \cline{3-14} 
 &  & DPC \cite{DPC} & 86.8 & 61.3 & - & - & - & - & - & - & - & 59.2 & - \\ \cline{3-14} 
 &  & InterpCNN \cite{Mao2019InterpConv} & - & - & 88.7 & 66.7 & - & - & - & - & - & - & - \\ \cline{2-14} 
 & \multirow{3}{*}{\begin{tabular}[c]{@{}c@{}}RNN \\-based\end{tabular}} & RSNet \cite{RSNet} & - & 51.9 & - & 56.5 & - & - & - & - & 84.9 & 39.4 & - \\ \cline{3-14} 
 &  & G+RCU \cite{engelmann2017exploring} & - & 45.1 & 81.1 & 49.7 & - & - & - & - & - & - & - \\ \cline{3-14} 
 &  & 3P-RNN \cite{3P-RNN} & 85.7 & 53.4 & 86.9 & 56.3 & - & - & - & - & - & - & - \\ \cline{2-14} 
 & \multirow{9}{*}{\begin{tabular}[c]{@{}c@{}}Graph\\-based\end{tabular}} & DGCNN \cite{Wang2019} & - & - & 84.1 & 56.1 & - & - & - & - & - & - & - \\ \cline{3-14} 
 &  & SPG \cite{SPGraph} & 86.4 & 58.0 & 85.5 & 62.1 & 92.9 & 76.2 & 94.0 & 73.2 & - & - & 17.4 \\ \cline{3-14} 
 &  & SSP+SPG \cite{oversegmentation} & 87.9 & 61.7 & 87.9 & 68.4 & - & - & - & - & - & - & - \\ \cline{3-14} 
 &  & GACNet \cite{GACNet} & 87.8 & 62.9 & - & - & - & - & 91.9 & 70.8 & - & - & - \\ \cline{3-14} 
 &  & PAG \cite{pointatrousgraph} & 86.8 & 59.3 & 88.1 & 65.9 & - & - & - & - & - & - &  \\ \cline{3-14} 
 &  & HDGCN \cite{HDGCN} & - & 59.3 & - & 66.9 & - & - & - & - & - & - & - \\ \cline{3-14} 
 &  & HPEIN \cite{HPEIN} & 87.2 & 61.9 & 88.2 & 67.8 & - & - & - & - & - & 61.8 & - \\ \cline{3-14} 
 &  & SPH3D-GCN \cite{Lei2018} & 87.7 & 59.5 & 88.6 & 68.9 & - & - & - & - & - & 61.0 & - \\ \cline{3-14} 
 &  & DPAM \cite{Liu2019DPAM} & 86.1 & 60.0 & 87.6 & 64.5 & - & - & - & - & - & - & - \\ \hline
\end{tabular}%
}
\end{table*}

\textbf{Point Convolution Methods.} 
These methods tend to propose effective convolution operators for point clouds. Hua et al. \cite{Hua2018} proposed a point-wise convolution operator, where the neighboring points are binned into kernel cells and then convolved with kernel weights. As shown in Fig. \ref{fig:Pointnet++}(b), Wang et al. \cite{PCCN} proposed a network called PCCN based on parametric continuous convolution layers. The kernel function of this layer is parameterized by MLPs and spans the continuous vector space. \why{Thomas} et al. \cite{kpconv} proposed a Kernel Point Fully Convolutional Network (KP-FCNN) based on Kernel Point Convolution (KPConv). Specifically, the convolution weights of KPConv are determined by the Euclidean distances to kernel points, and the number of kernel points is not fixed. The positions of the kernel points are formulated as an optimization problem of best coverage in a sphere space. Note that, the radius neighbourhood is used to keep a consistent receptive field, while grid subsampling is used in each layer to achieve high robustness under varying densities of point clouds. In \cite{DPC}, \why{Engelmann} et al. provided rich ablation experiments and visualization results to show the impact of receptive field on the performance of aggregation-based methods. They also proposed a Dilated Point Convolution (DPC) operation to aggregate dilated neighboring features, instead of the $\textit{K}$ nearest neighbours. This operation is demonstrated to be very effective in increasing the receptive field and can be easily integrated into existing aggregation-based networks. 

\textbf{RNN-based Methods.} To capture inherent context features from point clouds, Recurrent Neural Networks (RNN) have also been used for semantic segmentation of point clouds. Based on PointNet \cite{PointNet}, \why{Engelmann} et al. \cite{engelmann2017exploring} first transformed a block of points into multi-scale blocks and grid blocks to obtain input-level context. Then, the block-wise features extracted by PointNet are sequentially fed into Consolidation Units (CU) or Recurrent Consolidation Units (RCU) to obtain output-level context. Experimental results show that incorporating spatial context is important for the improvement of the segmentation performance. Huang et al. \cite{RSNet} proposed a lightweight local dependency modeling module, and utilized a slice pooling layer to convert unordered point feature sets into an ordered sequence of feature vectors. As shown in Fig. \ref{fig:Pointnet++}(c), Ye et al. \cite{3P-RNN} first proposed a Pointwise Pyramid Pooling (3P) module to capture the coarse-to-fine local structure, and then utilized two-direction hierarchical RNNs to further obtain long-range spatial dependencies. RNN is then applied to achieve an end-to-end learning. However, these methods lose rich geometric features and density distribution from point clouds when aggregating the local neighbourhood features with global structure features \cite{DarNet}. To alleviate the problems caused by the rigid and static pooling operations, Zhao et al. \cite{DarNet} proposed a Dynamic Aggregation Network (DAR-Net) to consider both global scene complexity and local geometric features. The inter-medium features are dynamically aggregated using a self-adapted receptive field and node weights. Liu et al. \cite{3DCNN-DQN-RNN} proposed 3DCNN-DQN-RNN for efficient semantic parsing of large-scale point clouds. This network first learns the spatial distribution and color features using a 3D CNN network, DQN is further used to localize objects belonging to a specific class. The final concatenated feature vector is fed into a residual RNN to obtain the final segmentation results.

\textbf{Graph-based Methods.} 
To capture the underlying shapes and geometric structures of 3D point clouds, several methods resort to graph networks. As shown in Fig. \ref{fig:Pointnet++}(d), \why{Landrieu} et al. \cite{SPGraph} represented a point cloud as a set of interconnected simple shapes and superpoints, and used an attributed directed graph (i.e., superpoint graph) to capture the structure and context information. Then, the large-scale point cloud segmentation problem is spilt into three sub-problems, i.e., geometrically homogeneous partition, superpoint embedding, and contextual segmentation. To further improve the partition step, \why{Landrieu and Boussaha} \cite{oversegmentation} proposed a supervised framework to oversegment a point cloud into pure superpoints. This problem is formulated as a deep metric learning problem structured by an adjacency graph. In addition, a graph-structured contrastive loss is also proposed to help the recognition of borders between objects. 

To better capture the local geometric relationships in high-dimensional space, Kang et al. \cite{pyramnet} proposed a PyramNet based on Graph Embedding Module (GEM) and Pyramid Attention Network (PAN). The GEM module formulates a point cloud as a directed acyclic graph and utilzes a covariance matrix to replace the Euclidean distance for the construction of adjacent similarity matrix. Convolution kernels with four different sizes are used in the PAN module to extract features with different semantic intensities. In \cite{GACNet}, Graph Attention Convolution (GAC) is proposed to selectively learn relevant features from a local neighboring set. This operation is achieved by dynamically assigning attention weights to different neighboring points and feature channels based on their spatial positions and feature differences. GAC can learn to capture discriminative features for segmentation, and has similar characteristics to the commonly used CRF model.
\Gary{Ma et al. \cite{ma2020global} proposed a Point Global Context Reasoning (PointGCR) module to capture global contextual information along the channel dimension using an undirected graph representation. PointGCR is a plug-and-play and end-to-end trainable module. It can easily be integrated into an existing segmentation network to achieve performance improvement.}

\qy{In addition, several very recent work tries to achieve semantic segmentation of point clouds under weak supervision. Wei et al. \cite{multipath} proposed a two-stage approach to train a segmentation network with subcloud level labels. Xu et al. \cite{xu2020weakly} investigated several inexact supervision schemes for  semantic segmentation of point clouds. They also proposed a network that is able to be trained with only partially labeled points (e.g. 10\%). }

\subsection{Instance Segmentation}
\begin{figure}[thb]
\centering
\includegraphics[width=0.5\textwidth]{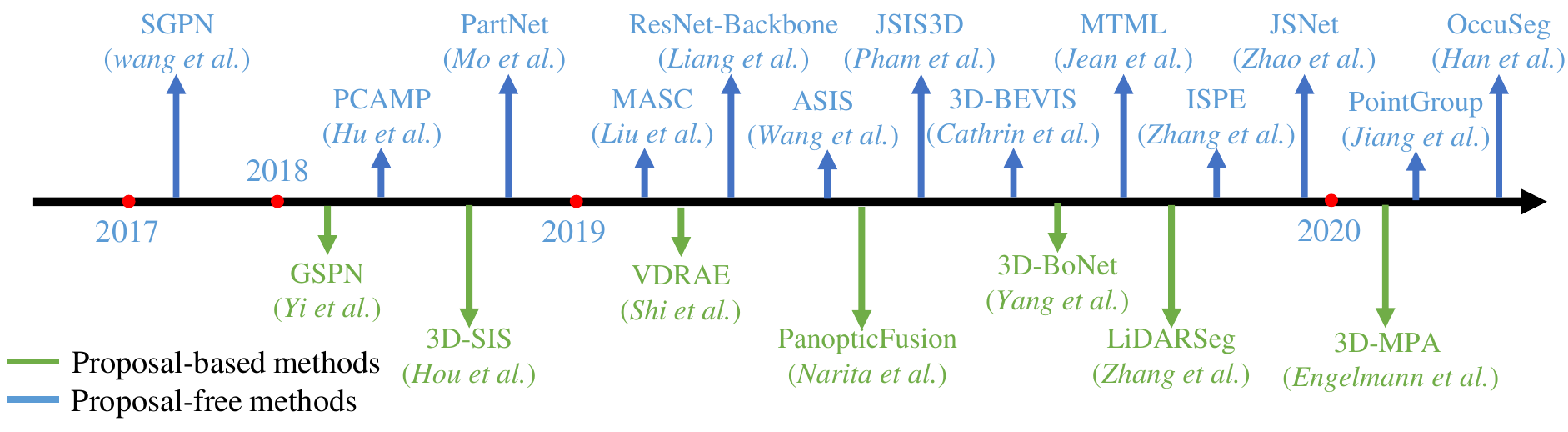}
\caption{Chronological overview of the most relevant deep learning-based 3D instance segmentation methods. }
\label{fig:milestoneins}
\end{figure}

Compared to semantic segmentation, instance segmentation is more challenging  as it requires more accurate and fine-grained reasoning of points. In particular, it not only needs to distinguish the points with different semantic meanings, but also separate instances with the same semantic meaning. Overall, existing methods can be divided into two groups: proposal-based methods and proposal-free methods. Several milestone methods are illustrated in Fig \ref{fig:milestoneins}.

\subsubsection{\bf{Proposal-based Methods}} 
These methods convert the instance segmentation problem into two sub-tasks: 3D object detection and instance mask prediction. 

Hou et al. \cite{3dsis} proposed a 3D fully-convolutional Semantic Instance Segmentation (3D-SIS) network to achieve semantic instance segmentation on RGB-D scans. This network learns from both color and geometry features. Similar to 3D object detection, a 3D Region Proposal Network (3D-RPN) and a 3D Region of Interesting (3D-RoI) layer are used to predict bounding box locations, object class labels and instance masks. Following the analysis-by-synthesis strategy, Yi et al. \cite{GSPN} proposed a Generative Shape Proposal Network (GSPN) to generate high-objectness 3D proposals. These proposals are further refined by a Region-based PointNet (R-PointNet). The final label is obtained by predicting a per-point binary mask for each class label. Different from direct regression of 3D bounding boxes from point clouds, this method removes a large amount of meaningless proposals by enforcing geometric understanding. 

By extending 2D panoptic segmentation to 3D mapping, \why{Narita} et al. \cite{panopticfusion} proposed an online volumetric 3D mapping system to jointly achieve large-scale 3D reconstruction, semantic labeling, and instance segmentation. They first utilized 2D semantic and instance segmentation networks to obtain pixel-wise panoptic labels and then integrated these labels to the volumtric map. A fully-connected CRF is further used to achieve accurate segmentation. This semantic mapping system can achieve high-quality semantic mapping and discriminative object recognition. Yang et al. \cite{3DBoNet} proposed a single-stage, anchor-free and end-to-end trainable network called 3D-BoNet to achieve instance segmentation on point clouds. This method directly regresses rough 3D bounding boxes for all potential instances, and then utilizes a point-level binary classifier to obtain instance labels. Particularly, the bounding box generation task is formulated as an optimal assignment problem. In addition, a multi-criteria loss function is also proposed to regularize the generated bounding boxes. This method does not need any post-processing and is computationally efficient. Zhang et al. \cite{feihu} proposed a network for instance segmentation of large-scale outdoor LiDAR point clouds. This method learns a feature representation on the bird's-eye view of point clouds using self-attention blocks. The final instance labels are obtained based on the predicted horizontal center and the height limits. Shi et al. \cite{shi2019hierarchy} proposed a hierarchy-aware Variational Denoising Recursive AutoEncoder (VDRAE) to predict the layout of indoor 3D space. The object proposals are iteratively generated and refined by recursive context aggregation and propagation.

Overall, proposal-based methods \cite{GSPN, 3dsis, 3DBoNet, 3D-MPA} are intuitive and straightforward, and the instance segmentation results usually have good objectness. However, these methods require multi-stage training and pruning of redundant proposals. Therefore, they are usually time-consuming and computationally expensive. 

\subsubsection{\bf{Proposal-free Methods}}  
Proposal-free methods \cite{SGPN, ASIS, jsis3d, 3DBEVIS, MASC, ResNetbackbone, Occuseg, jiang2020pointgroup} do not have an object detection module. Instead, they usually consider instance segmentation as a subsequent clustering step after semantic segmentation. In particular, most existing methods are based on the assumption that points belonging to the same instance should have very similar features. Therefore, these methods mainly focus on discriminative feature learning and point grouping.

In a pioneering work, Wang et al. \cite{SGPN} first introduced a Similarity Group Proposal Network (SGPN). This method first learns a feature and semantic map for each point, and then introduces a similarity matrix to represent the similarity between each paired features. To learn more discriminative features, they use a double-hinge loss to mutually adjust the similarity matrix and semantic segmentation results. Finally, a heuristic and non-maximal suppression method is adopted to merge similar points into instances. 
Since the construction of a similarity matrix requires large memory consumption, the scalability of this method is limited.
Similarly, Liu et al. \cite{MASC} first leveraged submanifold sparse convolution \cite{sparseconv} to predict semantic scores of each voxel and affinity between neighboring voxels. They then introduced a clustering algorithm to group points into instances based on the predicted affinity and the mesh topology. Mo et al. \cite{mo2019partnet} introduced a detection-by-segmentation network in PartNet to achieve instance segmentation. PointNet++ is used as the backbone to predict semantic labels of each point and disjoint instance masks.
Further, Liang et al. \cite{ResNetbackbone} proposed a structure-aware loss for the learning of discriminative embeddings. This loss considers both the similarity of features and the geometric relations among points. An attention-based graph CNN is further used to adaptively refine the learned features by aggregating different information from neighbors. 

Since the semantic category and instance label of a point are usually dependent on each other, several methods have been proposed to couple these two tasks into a single task.
Wang et al. \cite{ASIS} integrated these two tasks by introducing an end-to-end and learnable Associatively Segmenting Instances and Semantics (ASIS) module. Experiments show that semantic features and instance features can mutually support each other to achieve an improved performance through this ASIS module. Similarly, Zhao et al. \cite{zhao2019jsnet} proposed JSNet to achieve both semantic and instance segmentation. Further, Pham et al. \cite{jsis3d} first introduced a Multi-Task Point-wise Network (MT-PNet) to assign a label to each point and regularized the embeddings in the feature space by introducing a discriminative loss \cite{discriminative_loss}. They then fused the predicted semantic labels and embeddings to a Multi-Value Conditional Random Field (MV-CRF) model for joint optimization. Finally, mean-field variational inference is used to produce semantic labels and instance labels. Hu et al. \cite{hu2018semantic} first proposed a Dynamic Region Growing (DRG) method to dynamically separate a point cloud into a set of disjoint patches, and then used an unsupervised K-means++ algorithm to group all these patches. Multi-scale patch segmentation is then performed with the guidance of contextual information between patches. Finally, these labeled patches are merged into object level to obtain final semantic and instance labels.

To achieve instance segmentation on full 3D scenes,  \why{Elich} et al. \cite{3DBEVIS} presented a hybrid 2D-3D network to jointly learn global consistent instance features from a BEV representation and local geometric features of point clouds. The learned features are then combined to achieve semantic and instance segmentation. Note that, rather than heuristic \textit{GroupMerging} algorithms \cite{SGPN}, a more flexible Meanshift \cite{meanshift} algorithm is used to group these points into instances.
Alternatively, multi-task learning is also introduced for instance segmentation. \why{Lahoud} et al. \cite{MTML} learned both the unique feature embedding of each instance and the directional information to estimate the object's center. Feature embedding loss and directional loss are proposed to adjust the learned feature embeddings in latent feature space. Mean-shift clustering and non-maximum suppression are adopted to group voxels into instances. This method achieves the state-of-the-art performance on the ScanNet \cite{ScanNet} benchmark. Besides, the predicted directional information is particularly useful to determine the boundary of instances. Zhang et al. \cite{Probabilistic} introduced probabilistic embeddings to instance segmentation of point clouds. This method also incorporates uncertainty estimation and proposes a new loss function for the clustering step. \qy{Jiang et al. \cite{jiang2020pointgroup} proposed a PointGroup network, which is composed of a semantic segmentation branch and an offset prediction branch. A  dual-set clustering algorithm and the ScoreNet is further utilized to achieve better grouping results.}

In summary, proposal-free methods do not require computationally expensive region-proposal components. However, the objectness of instance segments grouped by these methods is usually low since these methods do not explicitly detect object boundaries.

\subsection{Part Segmentation}

The difficulties for part segmentation of 3D shapes are twofold. First, shape parts with the same semantic label have a large geometric variation and ambiguity. Second, the number of parts in objects with the same semantic meanings may be different.

VoxSegNet \cite{voxsegnet} is proposed to achieve fine-grained part segmentation on 3D voxelized data under a limited solution. A Spatial Dense Extraction (SDE) module (which consists of stacked atrous residual blocks) is proposed to extract multi-scale discriminative features from sparse volumetric data. The learned features are further re-weighted and fused by progressively applying an Attention Feature Aggregation (AFA) module. \why{Kalogerakis} et al. \cite{shapeFCN} combined FCNs and surface-based CRFs to achieve end-to-end 3D part segmentation. They first generated images from multiple views to achieve optimal surface coverage and fed these images into a 2D network to produce confidence maps. Then, these confidence maps are aggregated by a surface-based CRF, which is responsible for a consistent labeling of the entire scene. 
Yi et al. \cite{Syncspeccnn} introduced a Synchronized Spectral CNN (SyncSpecCNN) to perform convolution on irregular and non-isomorphic shape graphs. A spectral parameterization of dilated convolutional kernels and a spectral transformer network is introduced to solve the problem of multi-scale analysis in parts and information sharing across shapes. 

Wang et al. \cite{wang20183dSFCN} first performed shape segmentation on 3D meshes by introducing Shape Fully Convolutional Networks (SFCN) and taking three low-level geometric features as its input. They then utilized voting-based multi-label graph cuts to further refine the segmentation results. Zhu et al. \cite{cosegnet} proposed a weakly-supervised CoSegNet for 3D shape co-segmentation. This network takes a collection of unsegmented 3D point cloud shapes as input, and produces shape part labels by iteratively minimizing a group consistency loss. Similar to CRF, a pre-trained part-refinement network is proposed to further refine and denoise part proposals. Chen et al. \cite{BAENET} proposed a Branched AutoEncoder network (BAE-NET) for unsupervised, one-shot and weakly supervised 3D shape co-segmentation. This method formulates the shape co-segmentation task as a representation learning problem and aims at finding the simplest part representations by minimizing the shape reconstruction loss. Based on the encoder-decoder architecture, each branch of this network can learn a compact representation for a specific part shape. The features learned from each branch and the point coordinate are then fed to the decoder to produce a binary value (which indicates whether the point belongs to this part). This method has good generalization ability and can process large 3D shape collections (up to 5000+ shapes). However, it is sensitive to initial parameters and does not incorporate shape semantics into the network, which hinders this method to obtain a robust and stable estimation in each iteration. Yu et al. \cite{yu2019partnet} proposed a top-down recursive part decomposition network (PartNet) for hierarchical shape segmentation. Different from existing methods that segment a shape to a fixed label set, this network formulates part segmentation as a problem of cascade binary labeling, and decompose the input point cloud   to an arbitrary number of parts based on the geometric structure. Luo et al. \cite{group} introduced a learning-based grouping framework for the task of zero-shot 3D part segmentation. To improve the cross-category generalization ability, this method tends to learn a grouping policy that restricts the network to learn part-level features within the part local context.

\subsection{Summary}
Table \ref{tab:segmentation_results} shows the results achieved by existing methods on public benchmark, including S3DIS \cite{S3DIS}, Semantic3D \cite{Semantic3d}, ScanNet \cite{Scanet}, and SemanticKITTI \cite{semantickitti}. The following issues need to be further investigated:

\begin{enumerate}
    \item[$\bullet$] \qy{Thanks to the regular data representation, both  projection-based methods and discretization-based methods can leverage the mature network architecture from their 2D image counterparts. However, the main limitation of projection-based methods lies in the information loss caused by 3D-2D projection, while the main bottleneck for discretization-based methods is the cubically increased computational and memory costs caused by the increase of the resolution. To this end, sparse convolution building upon indexing structures would be a feasible solution and worth further exploration.}
    \item[$\bullet$] \qy{Point-based networks are the most frequently investigated methods. However, point representation naturally does not have explicit neighboring information, most existing point-based methods resort to expensive neighbor searching mechanisms  (e.g., KNN \cite{li2018pointcnn} or ball query \cite{PointNet++}). This inherently limits the efficiency of these methods, the recently proposed point-voxel joint representation \cite{point-voxel} would be an interesting direction for further investigation.}
    
   \item[$\bullet$]  Learning from imbalanced data is still a challenging problem in point cloud segmentation. Although several approaches \cite{zhang2019shellnet, kpconv, SPGraph} have achieved a remarkable overall performance, their performance on minority classes is still limited. For example, RandLA-Net \cite{randla} achieves an overall IoU of 76.0\% on the \textit{reduced-8} subset of Semantic3D, but a very low IOU of 41.1\% on the class of \textit{hardscape}.
   
   \item[$\bullet$] The majority of existing approaches \cite{PointNet, PointNet++, li2018pointcnn, LSANet, zhang2019shellnet} work on small point clouds (e.g., 1m$\times$1m with 4096 points). In practice, the point clouds acquired by depth sensors are usually immense and large-scale. Therefore, it is desirable to further investigate the problem of efficient segmentation of large-scale point clouds.
   
   \item[$\bullet$] A handful of works \cite{fan2019pointrnn, MeteorNet, Minkowski} have started to learn spatio-temporal information from dynamic point clouds. It is expected that the spatio-temporal information can help to improve the performance of subsequent tasks such as 3D object recognition, segmentation, and completion. 

\end{enumerate}

\section{Conclusion} \label{sec:conclusion}
This paper has presented a contemporary survey of the state-of-the-art methods for 3D understanding, including 3D shape classification, 3D object detection and tracking, and 3D scene and object segmentation. A comprehensive taxonomy and performance comparison of these methods have been presented. Merits and demerits of various methods are also covered, with potential research directions being listed.

\ifCLASSOPTIONcompsoc
  \section*{Acknowledgments}
\else
  \section*{Acknowledgment}
\fi

This work was partially supported by the National Natural Science Foundation of China (No. 61972435, 61602499, 61872379), the Natural Science Foundation of Guangdong Province (2019A1515011271), the Science and Technology Innovation Committee of Shenzhen Municipality (JCYJ20190807152209394), the Australian Research Council (Grants DP150100294 and DP150104251), the China Scholarship Council (CSC) and the Academy of Finland.

\bibliographystyle{IEEEtran}
\bibliography{TPAMI_ref}

\ifCLASSOPTIONcaptionsoff
  \newpage
\fi

\begin{IEEEbiography}[{\includegraphics[width=1in,height=1.25in,clip,keepaspectratio]{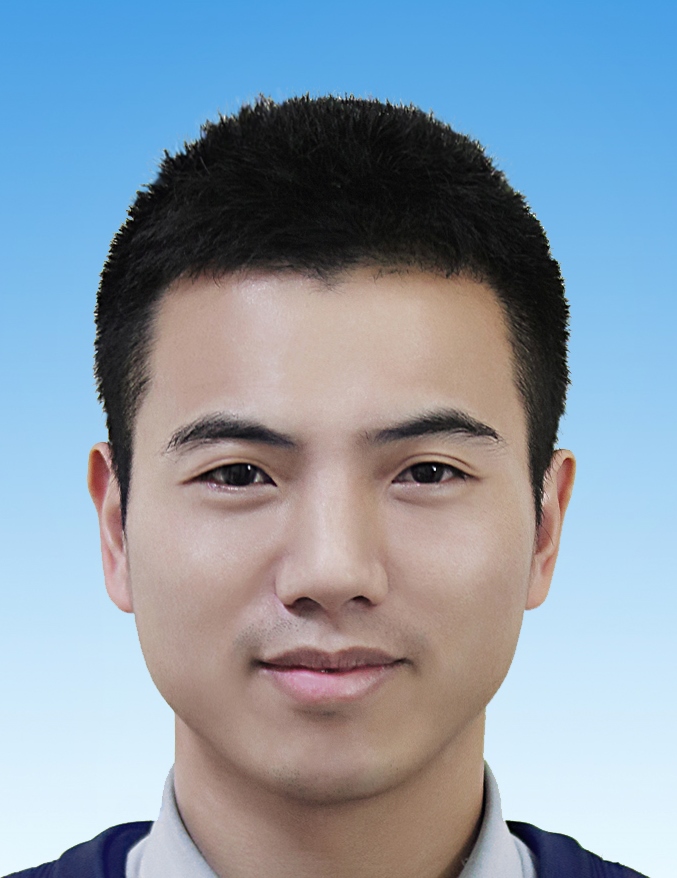}}]{Yulan Guo} is currently as associate professor. He received the B.Eng. and Ph.D. degrees from National University of Defense Technology (NUDT) in 2008 and 2015, respectively. He was a visiting Ph.D. student with the University of Western Australia from 2011 to 2014. He worked as a postdoctorial research fellow with the Institute of Computing Technology, Chinese Academy of Sciences from 2016 to 2018. He has authored over 90 articles in journals and conferences, such as the IEEE TPAMI and IJCV. His current research interests focus on 3D vision, particularly on 3D feature learning, 3D modeling, 3D object recognition, and scene understanding. Dr. Guo received the ACM China SIGAI Rising Star Award in 2019, Wu-Wenjun Outstanding AI Youth Award in 2019, and the CAAI Outstanding Doctoral Dissertation Award in 2016. He served as an associate editor for IET Computer Vision and IET Image Processing, a guest editor for IEEE TPAMI, and an area chair for CVPR 2021 and ICPR 2020.
\end{IEEEbiography}

\begin{IEEEbiography}[{\includegraphics[width=1in,height=1.25in,clip,keepaspectratio]{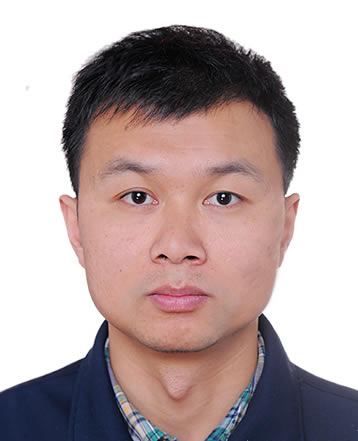}}]{Hanyun Wang} received his Ph.D. degree from National University of Defense Technology in 2015. He was a visiting Ph.D. student with Xiamen University from 2011 to 2014.  He has authored over 20 articles in journals and conferences, such as IEEE TGRS and IEEE TITS. His research interests include mobile laser scanning data analysis and 3D computer vision, especially on 3D object detection and 3D scene understanding. He also served as reviewers for many journals, such as IEEE TGRS, IEEE GRSL and IET Image Processing.
\end{IEEEbiography}

\begin{IEEEbiography}[{\includegraphics[width=1in,height=1.25in,clip,keepaspectratio]{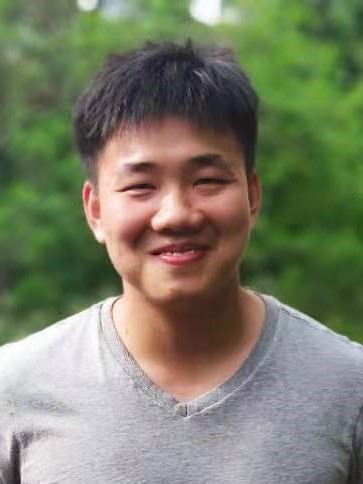}}]{Qingyong Hu} received his M.Eng. degree in information and communication engineering from the National University of Defense Technology (NUDT) in 2018. He is currently a DPhil candidate in the Department of Computer Science at the University of Oxford. His research interests lie in 3D computer vision, large-scale point cloud processing, and visual tracking.
\end{IEEEbiography}

\begin{IEEEbiography}[{\includegraphics[width=1in,height=1.25in,clip,keepaspectratio]{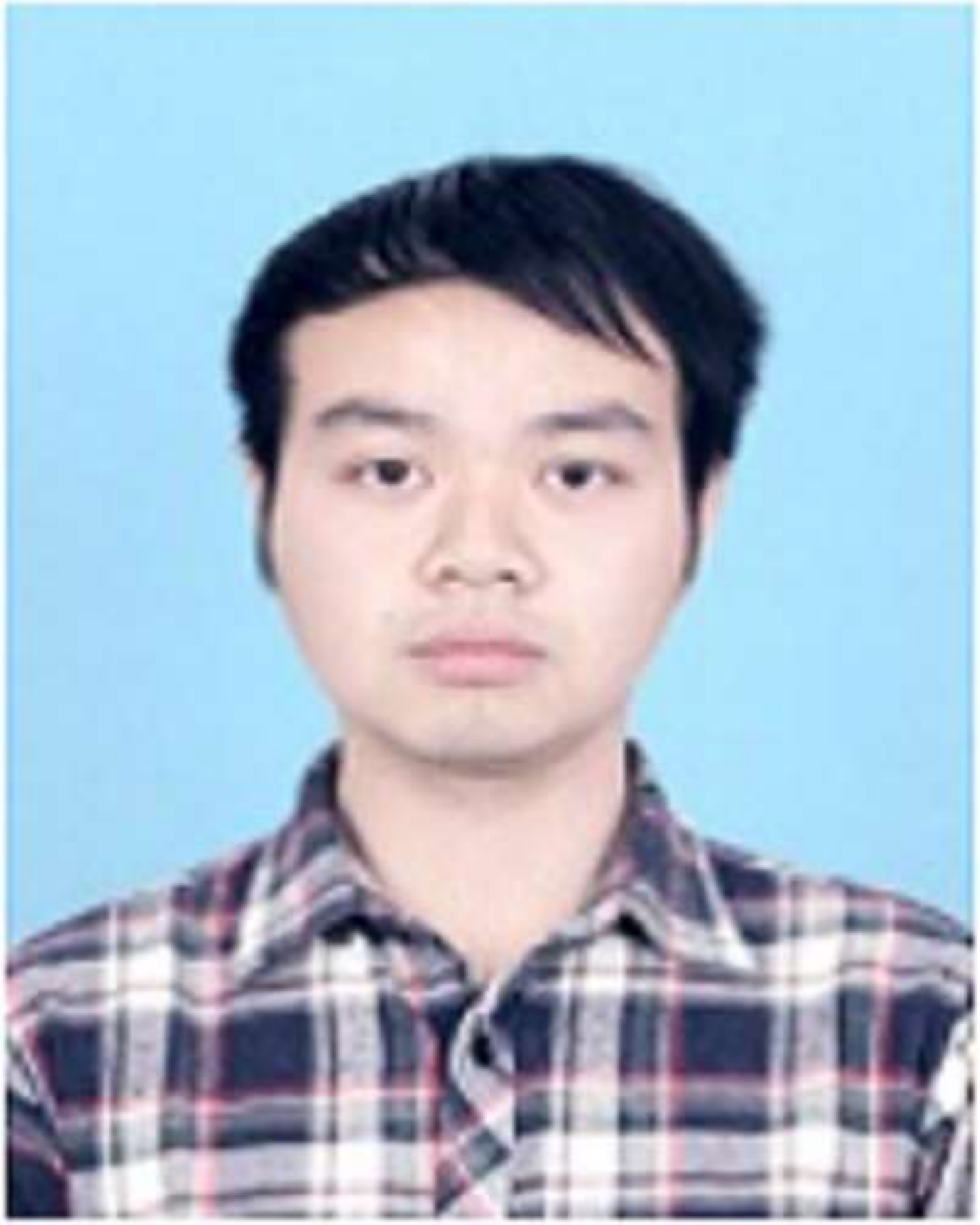}}]{Hao Liu} received the B.Eng. degree from University of Electronic Science and Technology of China (UESTC) in 2016, and M.S. degree from National University of Defense Technology (NUDT) in 2018. He is currently pursuing the Ph.D. degree with the School of Electronics and Communication Engineering, Sun Yat-sen University. His research interests lie in 3D computer vision and point cloud processing.
\end{IEEEbiography}

\begin{IEEEbiography}[{\includegraphics[width=1in,height=1.25in,clip]{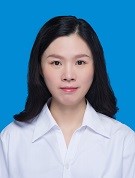}}]{Li Liu}
received the BSc degree in communication engineering, the MSc degree in photogrammetry and remote sensing and the Ph.D. degree in information and communication engineering from the National University of Defense Technology (NUDT), China, in 2003, 2005 and 2012, respectively. She joined the faculty at NUDT in 2012, where she is currently an Associate Professor with the College of System Engineering. During her PhD study, she spent more than two years as a Visiting Student at the University of Waterloo, Canada, from 2008 to 2010. From 2015 to 2016, she spent ten months visiting the Multimedia Laboratory at the Chinese University of Hong Kong. From 2016.12 to 2018.11, she worked as a senior researcher at the Machine Vision Group at the University of Oulu, Finland. She was a cochair of nine International Workshops at CVPR, ICCV, and ECCV. She was a guest editor of special issues for IEEE TPAMI and IJCV. Her current research interests include computer vision, pattern recognition and machine learning. Her papers have currently over 2300+ citations in Google Scholar. She currently serves as Associate Editor of the Visual Computer Journal and Pattern Recognition Letter. She serves as Area Chair of ICME 2020.\end{IEEEbiography}

\begin{IEEEbiography}[{\includegraphics[width=1in,height=1.25in,clip]{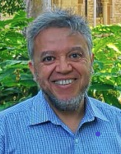}}]{Mohammed Bennamoun} is Winthrop Professor in the Department of Computer Science and Software Engineering at UWA and is a researcher in computer vision, machine/deep learning, robotics, and signal/speech processing. He has published 4 books (available on Amazon), 1 edited book, 1 Encyclopedia article, 14 book chapters, 120+ journal papers, 250+ conference publications, 16 invited \& keynote publications. His h-index is 50 and his number of citations is 11,000+ (Google Scholar). He was awarded 65+ competitive research grants, from the Australian Research Council, and numerous other Government, UWA and industry Research Grants. He successfully supervised 26+ PhD students to completion. He won the Best Supervisor of the Year Award at QUT (1998), and received award for research supervision at UWA (2008 \& 2016) and Vice-Chancellor Award for mentorship (2016).  He delivered conference tutorials at major conferences, including: IEEE Computer Vision and Pattern Recognition (CVPR 2016), Interspeech 2014, IEEE International Conference on Acoustics Speech and Signal Processing (ICASSP) and European Conference on Computer Vision (ECCV). He was also invited to give a Tutorial at an International Summer School on Deep Learning (DeepLearn 2017).
\end{IEEEbiography}




\end{document}